\documentclass[acmtog,authorversion,nonacm]{acmart}

\usepackage{booktabs} %

\usepackage[ruled]{algorithm2e} %

\SetAlFnt{\small}
\SetAlCapFnt{\small}
\SetAlCapNameFnt{\small}
\SetAlCapHSkip{0pt}

\copyrightyear{2023}
\acmYear{2023}
\setcopyright{acmlicensed}\acmConference[SA Conference Papers '23]{SIGGRAPH Asia 2023 Conference Papers}{December 12--15, 2023}{Sydney, NSW, Australia}
\acmBooktitle{SIGGRAPH Asia 2023 Conference Papers (SA Conference Papers '23), December 12--15, 2023, Sydney, NSW, Australia}
\acmPrice{15.00}
\acmDOI{10.1145/3610548.3618156}
\acmISBN{979-8-4007-0315-7/23/12}

\usepackage{color}

\newcommand{\figref}[1]{Fig.~\ref{#1}}
\newcommand{\tabref}[1]{Table~\ref{#1}}
\newcommand{\eqnref}[1]{Eq.~\ref{#1}}
\newcommand{\secref}[1]{Section~\ref{#1}}

\definecolor{dbcolor}{RGB}{0,150,200}
\definecolor{lccolor}{RGB}{10,10,210}
\definecolor{pccolor}{RGB}{10,210,10}
\definecolor{escolor}{RGB}{210,10,10}
\definecolor{pgcolor}{RGB}{210,10,210}

\newcommand\review[1] {#1}

\newcommand{\remove}[1]{{}}
\newcommand{\ie}{{i.e.}}
\newcommand{\eg}{{\em e.g.}}

\newcommand{\SO}[0]{\mathrm{SO}}

\newcommand{\z}[0]{\mathbf{z}}
\newcommand{\s}[0]{\mathbf{s}}              %
\renewcommand{\u}[0]{\mathbf{u}}            %
\newcommand{\X}[0]{\mathbf{X}}              %
\newcommand{\x}[0]{\mathbf{x}}              %
\newcommand{\y}[0]{\mathbf{y}}              %
\newcommand{\A}[0]{\mathbf{A}}              %
\newcommand{\F}[0]{\mathbf{F}}              %
\newcommand{\R}[0]{\mathbf{R}}              %
\newcommand{\W}[0]{\mathbf{W}}              %
\newcommand{\network}[0]{\mathcal{N}}             %
\newcommand{\mesh}[0]{\mathbf{T}}

\newcommand{\densityE}[0]{\Psi}
\newcommand{\domain}[1]{\mathcal{D}_{#1}}

\usepackage{tabularx}

\newcolumntype{P}[1]{>{\centering\arraybackslash}p{#1}}

\usepackage[normalem]{ulem}
\usepackage{stackengine}
\usepackage{bm}
\usepackage[ruled]{algorithm2e} %
\usepackage{wrapfig}
\usepackage[export]{adjustbox}
\usepackage{tabularx}
\usepackage{tikz}
\usepackage{booktabs}
\usepackage{graphicx}
\usepackage{subcaption} 
\usepackage{tikz}
\usepackage{rotating}
\usepackage{adjustbox}
\usepackage{multirow}
\usepackage{multicol}

\begin{document}
\title{An Implicit Physical Face Model Driven by Expression and Style}

\author{Lingchen Yang}
\orcid{0000-0001-9918-8055}
\affiliation{\institution{ETH Zurich}\country{Switzerland}}
\email{lingchen.yang@inf.ethz.ch}

\author{Gaspard Zoss}
\orcid{0000-0002-0022-8203}
\affiliation{\institution{DisneyResearch|Studios}\country{Switzerland}}
\email{gaspard.zoss@disneyresearch.com}

\author{Prashanth Chandran}
\orcid{0000-0001-6821-5815}
\affiliation{\institution{DisneyResearch|Studios}\country{Switzerland}}
\email{prashanth.chandran@disneyresearch.com}

\author{Paulo Gotardo}
\orcid{0000-0001-8217-5848}
\affiliation{\institution{DisneyResearch|Studios}\country{Switzerland}}
\email{gotardop@gmail.com}

\author{Markus Gross}
\orcid{0009-0003-9324-779X}
\affiliation{\institution{ETH Zurich}\country{Switzerland}}
\affiliation{\institution{DisneyResearch|Studios}\country{Switzerland}}
\email{gross@disneyresearch.com}

\author{Barbara Solenthaler}
\orcid{0000-0001-7494-8660}
\affiliation{\institution{ETH Zurich}\country{Switzerland}}
\email{solenthaler@inf.ethz.ch}

\author{Eftychios Sifakis}
\orcid{0000-0001-5608-3085}
\affiliation{\institution{University of Wisconsin Madison}\country{USA}}
\email{sifakis@cs.wisc.edu}

\author{Derek Bradley}
\orcid{0000-0002-2055-9325}
\affiliation{\institution{DisneyResearch|Studios}\country{Switzerland}}
\email{derek.bradley@disneyresearch.com}

\renewcommand\shortauthors{L. Yang, G. Zoss, P. Chandran, P. Gotardo, M. Gross, B. Solenthaler, E. Sifakis, D. Bradley}

\begin{abstract}

3D facial animation is often produced by manipulating facial deformation models (or rigs), that are traditionally parameterized by expression controls.  A key component that is usually overlooked is expression ``style", as in, {\em how} a particular expression is performed. Although it is common to define a semantic basis of expressions that characters can perform, most characters perform each expression in their own style.  To date, style is usually entangled with the expression, and it is not possible to transfer the style of one character to another when considering facial animation.  We present a new face model, based on a data-driven implicit neural physics model, that can be driven by both expression and style separately.  At the core, we present a framework for learning implicit physics-based actuations for multiple subjects simultaneously, trained on a few arbitrary performance capture sequences from a small set of identities.  Once trained, our method allows generalized physics-based facial animation for any of the trained identities, extending to unseen performances. Furthermore, it grants control over the animation style, enabling style transfer from one character to another or blending styles of different characters. Lastly, as a physics-based model, it is capable of synthesizing physical effects, such as collision handling, setting our method apart from conventional approaches.
Please refer to our \href{https://studios.disneyresearch.com/2023/11/29/an-implicit-physical-face-model-driven-by-expression-and-style/}{{\textcolor{orange}{\emph{project page}}}} for more details.

\end{abstract}



 \begin{teaserfigure}
	\begin{center}
 	\includegraphics[width=1.0\linewidth]{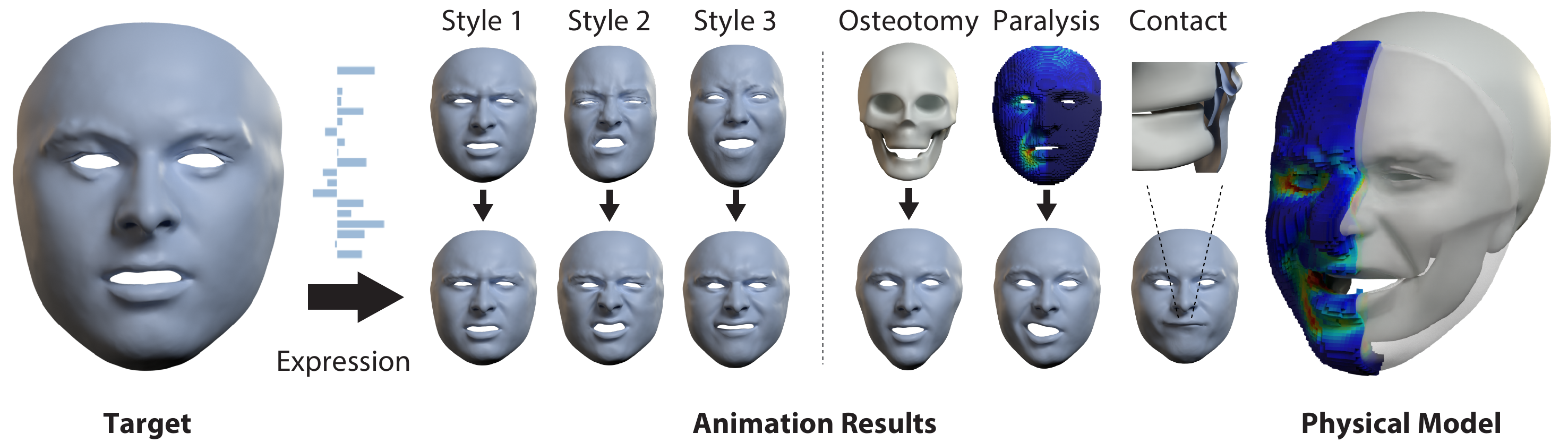}
 	\caption{Our physical face model, learned from multi-identity data, is driven by expression and style, allowing to synthesize different styles for the same expression. In addition, it supports various physical effects, such as bone reshaping (osteotomy), paralysis, contact and collision handling, etc.}
 	\label{fig:teaser}
	\end{center}
 \end{teaserfigure}

\maketitle

\section{Introduction}
\label{sec:intro}

Facial animation is a key component in computer graphics, especially for video games, visual effects, telepresence and beyond. 3D faces are generally controlled by facial rigs, which offer intuitive parameters to control the facial deformation.  A common example is blendshape-based animation~\cite{lewis2014egstarblendshape}, where faces are created by combining weighted contributions of different blendshapes.  Here, the blendweight vector defines the {\em expression} of the character.  Animation is defined by a sequence of blendweight vectors, often called animation curves, which can be key-framed by artists or reconstructed from motion-capture data.  Given different characters with semantically corresponding blendshape rigs, the same blendweights can be applied to each rig in order to obtain the same expression but performed by the different characters, in their own style, defined by each character's unique blendshapes.  

An interesting component of facial animation that is largely overlooked is the individual {\em style} of performing expressions.  For example, if you ask two different people to perform a large smile (which is often one of the basis blendshapes), undoubtedly the two people will activate different muscles with different intensities, depending on the particular individual style. This style component is usually baked-in to the blendshapes for each character, and cannot be easily manipulated by artists. In this work, we propose a new facial animation model that allows control over both expression and style.

As we hinted at, individual styles can be characterized by different muscle activations that one character may perform as compared to another. This suggests the use of a physics-based actuation model for facial animation.
Several physical face models have been proposed, evolving from the initial, labor-intensive first-principles muscle models~\cite{Sifakis05} to very recent and innovative implicit models that \emph{obviate} the need for precise anatomical structures~\cite{yang2022implicit}.  While the latter approach seems attractive for our application, the model by Yang et al. is designed for a single identity, and a completely new model would need to be trained for each individual muscle actuation style. 
In order to understand and represent the nuances in actuation styles across identities, we need to be able to analyze multiple identities simultaneously, within the same physical model.  
As such, we propose a novel implicit physical face model that is concurrently trained on animation data from multiple identities. By incorporating a broader range of human expression styles, we aim to create a more inclusive, comprehensive, and holistic system for simulating facial actuations.

For this goal, we make several informed design decisions.  Our initial observation centers on the use of an implicit neural network, which is agnostic to anatomy discretization. This feature allows us to bypass the laborious process of building topologically-consistent simulation meshes across different identities, a process that often grapples with the trade-off between skin-conforming accuracy and element quality. Consequently, we enable each identity to have its own precise simulation mesh with well-sized elements. Second, in observing that each individual possesses a unique material space defining their actuation field, we recognize that directly learning our model within these spaces neglects the shared muscle structures common among individuals. Consequently, to foster a more unified muscle structure, we propose a canonical space to let our model learn from such regularities. Thus, each identity-specific material space is continuously mapped to this shared space, separating the learning and simulation spaces.  Third, 
we parameterize expressions using traditional blendshape weights for straightforward control. Meanwhile, style is parameterized by quantized style codes, one for each identity.
Our network is trained using performance data from a small selection of identities.  In addition, we show how Lipschitz weight regularization, previously used only in geometry generation \cite{liu2022learning}, can be used to help smooth actuation generation, and, more importantly, disentangle the expression and style spaces. Lastly, a robust and versatile collision model is integrated.

Allowing multi-identity training for an implicit physics-based model opens up several interesting possibilities that have not been demonstrated before.  For example, we show that it is possible to learn generalized actuations across the identities.  As such, we do not require the different identities to perform the same expressions for training, and can generalize to new animations not seen at training time.  Furthermore, as a key innovation of our method, multi-identity training allows us to learn identity-specific styles and interpolate between them at inference time.  These properties offer distinct advantages over traditional facial rigs, which we demonstrate across animation tasks such as style transfer and retargeting.  Finally, we manifest that our physics-based model allows physical effects like contact and collision handling, manipulating the anatomy (e.g. resizing bones), the addition of extra forces (such as friction), and artistic control of muscle activation strength (e.g. to generate realistic facial paralysis).  These applications highlight the superiority of our method, setting it apart from conventional approaches.

\section{Related Work}
\label{sec:related}

We discuss related research in the areas of physics-based facial animation, facial animation retargeting and animation style transfer.
\review{For a more comprehensive overview, we refer to Egger et al.~\shortcite{egger20203d}}.

\subsection{Physics-Based Facial Animation}

Physics-based methodologies have been employed to animate faces, aiming to capture complex, nonlinear deformations more accurately~\cite{Sifakis05}. This often involves a volumetric discretization process that incorporates anatomical features, such as bones and muscles, to create a more lifelike representation. The physics simulators can then model intricate effects, including internal tissue and bone contacts and interactions with external objects. 
This has been demonstrated, for example, in the simulation of temporal dynamics due to inertia and external forces~\cite{ichim2017phace}, mandible kinematics~\cite{dorda2022Orthognathic}, or lip sliding~\cite{kim2019LipSliding}. More recently, physics simulators have also been used to learn anatomically inspired face models for animation and retargeting \cite{Choi2022}.
However, developing these simulation-ready facial meshes is laborious. Given this, there's been an inclination to merge physics with traditional blendshape models. This amalgamation attempts to bring together the best of both worlds: the simplicity of blendshapes and the enhanced realism of physics-based simulations~\cite{barielle2016Blendforces}. While these combined models do offer benefits, such as a physics layer to account for secondary movements and interactions~\cite{Ichim2016Blendvolumes}, they are inherently limited in terms of physical realism.

Recently, differentiable physics solvers have emerged as a popular tool in facial animation to solve complex inverse problems. These solvers, through techniques like the differentiable Finite Element Method, allow for the optimization of mechanical properties of the face, in particular, soft tissue parameters, heterogeneous stiffness and prestrain~\cite{Kadlecek2019BuildingData}, for enhanced realism.  They've also been applied in diverse scenarios, such as determining parameters for surgical treatments using a visible facial 3D surface~\cite{dorda2022Orthognathic} or in muscle-based systems for both shape matching and facial pose recognition from a single image input~\cite{Bao19}. Importantly, newer models are shifting away from requiring complex muscle structures. Instead, they focus on a shape-targeting approach~\cite{klar2020Shapetargeting}, where underlying elements of the mesh are all regarded as active and actuated to represent facial expressions. Srinivasan et al.~\shortcite{srinivasan2021learning} combined neural networks with this general muscle model, to learn the muscle actuation mechanism. Later studies moved towards an implicit neural representation, making this actuation mechanism more compact, adaptable and resolution-independent~\cite{yang2022implicit}.
Our proposed face model is similar to the one of Yang et al.~\shortcite{yang2022implicit}, but we train the network on captured data from multiple individuals simultaneously, which is key to learning different expression styles based on actuation. This multi-identity training comes with challenges regarding different style, geometry and simulation topology per identity, which we address in our method. 

\subsection{Facial Animation Retargeting}

Our implicit physical face model can be used for the application of facial animation retargeting.  The most commonly-used approach for retargeting is to employ blendshape-based facial rigs~\cite{lewis2014egstarblendshape}, where face deformations are formed by blending linear combinations of pre-defined semantic blendshapes.  If the source and target character rigs share the same semantics, retargeting is as simple as transferring the blendweights corresponding to an animation.  Several attempts have been made to improve on blendshape-based facial retargeting, such as adding supplemental blendshapes~\cite{Kim2011face}, coefficient re-mapping for characteristic blendshapes~\cite{Song2011}, calibrating the range of motion between source and target characters~\cite{Ribera2017}, and considering local models~\cite{liu2011framework,Xu2014,Chandran2022Karacast}.

Blendshape rigs are usually costly to create, due to the large number of facial shapes that are required for high quality, and retargeted results often require additional artist editing~\cite{Seol2011artist}.  On the other hand, animation retargeting can be achieved using a basic deformation transfer approach~\cite{Sumner2004DefTrans}, which requires only a single neutral expression of the target character.  This method is only suitable when all deformation details, including fine-scale wrinkles, should be transferred from the source to the target character. %
One limitation of many face models is that they are linear, limiting the expressiveness of motion. To counter this limitation, nonlinear deep face models have been proposed~\cite{Chandran2020,Chandran2022PerfSynth}. 
\review{Neural networks specifically designed for retargeting have also been proposed~\cite{Costigan2014,chaudhuri2019joint,bai2021riggable,Zhang2022,athar2022rignerf}}.

Other approaches to facial animation retargeting include expression cloning~\cite{Noh2001,Seol2012spacetime}, retargeting through a 2D image space~\cite{Kim2021deep, Moser2021}, dedicated lip-sync retargeting using contours~\cite{Bhat2013}, and real-time puppeteering~\cite{Li2013,Bouaziz2013,Cao2014}. The current state-of-the-art is the local, anatomically-constrained method of Chandran et al.~\shortcite{Chandran2022Karacast}, which combines local blendshape weight transfer with skin and bone sliding constraints.  All approaches described in this section, fail to consider the style of the expression during retargeting and do not offer animation style transfer.  Our approach supports facial animation retargeting, allowing retargeting of expression, style or both components from a source performance to a target character.  Furthermore, existing retargeting approaches do not consider physical effects like lip contact, collision with internal teeth and bone structures, or other physical effects that allow artist control over the retargeted performance.

\subsection{Animation Style Transfer}

A core component of our model is its ability to transfer animation style between characters.  Representing and transferring style has become very common in the domain of image processing~\cite{Huang2017,Jing2020} (refer to Jing et al.~\shortcite{Jing2019} for a detailed survey).  In terms of facial images and video, learning a particular speaker's motion style is a popular component of audio-based 2D facial animation~\cite{Suwajanakorn2017Obama,Zhou2020MakeItTalk,Pham2023StyleTransfer}.  This concept has been applied also in audio-driven 3D facial animation, where the goal is to create 3D animation from speech with individual style~\cite{Cudeiro2019,Lahiri2021LipSync3D,Richard2021meshtalk,Thambiraja2022Imitator}.

Full body motion style transfer for 3D skeleton animation has been a well-studied topic~\cite{Park2021,Tao2022,Jang2022MotionPuzzle}, including learning style from video input~\cite{Aberman2020unpaired}.  To our knowledge, facial animation style transfer has been largely overlooked in literature, and given today's tools, we believe it is extremely difficult to create animations that contain the geometry of one character but in the style of another character.  To address this challenge, we present the first implicit physics method trained on multiple identities, driven by both expression and style, allowing facial animation style transfer.

\section{Background}
\label{sec:background}

We first review the quasistatic, differentiable, active soft-body simulation and actuation networks, and then introduce our physics-based facial animation model.
We refer the readers to previous work \cite{srinivasan2021learning, yang2022implicit} for more details.

Human bodies and faces are examples of active objects that can change their shape through internal actuation mechanisms. 
In facial simulation, the actuation mechanism is modeled using the actuation tensor field $\mathcal{A}$, which is defined over the material space $\domain{M}$ (undeformed facial space) and consists of local $3 \times 3$ symmetric matrices $\A$ that contain information about the contractile directions and magnitudes. 
Given the measurement of local deformation with deformation gradient $\F \in \mathbb{R}^{3 \times 3}$, the shape targeting model \cite{klar2020Shapetargeting} is often used as the energy density function for simulation:
\begin{equation}\label{eq:shapetarget}
    \densityE(\F, \A) = \underset{\R \in \SO(3)}{\operatorname{min}}||\F - \R\A||^2_F.
\end{equation}
Here, the rotation matrix $\R$ factors away the rotational difference between $\F$ and $\A$ and makes $\densityE$ rotationally-invariant. The minimum of the total energy implicitly defines the sensitivity matrices of the simulation, which can be used to compute the gradient flowing into $\mathcal{A}$, enabling the training of generative actuation networks.

The influential work of Srinivasan et al.~\shortcite{srinivasan2021learning} proposes an explicit \emph{discretization-attached} actuation network replicating muscle actuation mechanisms and a latent activation space. 
Later, Yang et al.~\shortcite{yang2022implicit} advances this approach by modeling actuation as a continuous function in the material space $\domain{M}$, parameterized by a coordinate-based neural network $\network_{A}$, \ie, $\mathcal{A}(\x) = \network_{A}(\x; \z), \x \in \domain{M}$, for a sampled activation code $\z$ in the activation space. This eliminates the dependency on mesh discretization. The generated actuation field $\mathcal{A}$ is subsequently fed into the differentiable simulation to produce the deformed facial geometry. 
Such a model deviates from the purely geometry-based methods in that, after training, it generalizes better to novel and unseen physical effects. Besides, it requires no accurate knowledge of the underlying anatomy.

\section{Methodology}
\label{sec:method}

In this section, we present our novel method.
We build on top of the implicit, continuous actuation model of Yang et al.~\shortcite{yang2022implicit}, leveraging its discretization (topology) agnosticism, and propose several extensions to enable more efficient multi-identity learning, and thus controllability of both expression and individual style.

\subsection{Dataset and Pre-Processing}
\label{subsec:trainingData}
Our training data consists of 3D facial animation sequences of nine identities performing various actions such as dialogue and random facial expressions, sourced from the work of Chandran et al.~\shortcite{Chandran2022Karacast}. For each identity, we require a small amount of performance data, typically less than 30 seconds, and a collection of 20 static facial expressions. While these expressions are semantically consistent across identities, they embody unique styles, defining an individual blendshape basis for each identity. The performance sequences are projected into the respective blendshape basis of each identity, by optimizing for per-frame blendweight vectors, which our method will use as input.

The above collected data includes bone meshes, consisting of a skull and mandible, unique to each identity. 
For each frame, we also track the mandible position, utilizing the method introduced by Zoss et al.~\shortcite{Zoss2019}. Additionally, for every identity, we construct a simulation mesh. This mesh is discretized from the undeformed volumetric soft tissue between the facial surface and the bone, \ie, the material space, using regular finite hexahedral elements. We then embed the captured surface and bone meshes into the simulation mesh via trilinear interpolation, as shown in \figref{fig:teaser} (right).

The result of data processing is, for each identity, a sequence of facial surface meshes with bone geometry in \emph{topological correspondence} and corresponding blendshape weights, plus a single \emph{identity-specific} simulation mesh of hexahedral elements with facial geometry and bone attachments embedded.
We assign six identities for training, while the rest are reserved for cross-identity assessments such as retargeting. For numerical reconstruction evaluation, three sequences, one for each of the three identities from the training set, are specifically designated (approximately 10 seconds). 

\subsection{\review{Creation of Canonical Space and Mappings}}
Given that distinct identities each have their unique material spaces that define their actuation fields, we aim to develop a unified, aligned, and implicit actuation model applicable to all identities. 
For this, we introduce a canonical space $\domain{C}$, \review{which is a 3D volume that contains the canonical facial surface and bone meshes, sculpted by the artist.}  Furthermore, we establish a mapping function $\phi^{id}(\cdot)$ per identity, transforming identity-specific material space $\domain{M}^{id}$ into $\domain{C}$. 
This separates simulation and learning spaces, while still retaining a linkage between them. Concretely, the simulator first samples 3D spatial positions $\x \in \domain{M}^{id}$, which are subsequently warped as $\X \in \domain{C}$ by the mapping function $\phi^{id}$. These $\X$ positions are used to query the actuation network, resulting in the actuation tensors $\mathcal{A}(\X)$ required for the simulation. 
Consequently, the implicit actuation field, now defined on $\domain{C}$, becomes shared by different identities.

For each identity, we train a small MLP network to represent the mapping (deformation) \mbox{$\X = \phi^{id}(\x)$}. 
\review{We supervise the mapping using points on the surface meshes, thanks to the topological consistency.  For the rest of the volume off-surface, we add an elastic regularization term to smooth the volumetric mapping, similar to related work~\cite{park2021nerfies,wang2022morf}}. 
The total mapping loss for a single identity is defined as:
\begin{equation}
    \label{eq:mapping_loss}
\mathcal{L}_{map}^{id} =  \frac{1}{|\mesh_{C}|}||\phi^{id}(\mesh_{M}) - \mesh_{C}|| + \frac{\lambda_e}{N_e} \sum_{i=1}^{N_e} \underset{\R \in \SO(3)}{\operatorname{min}}||\nabla_{\x_i} \phi^{id}(\x_i) - \R||,
\end{equation}
where \review{$\mesh$ denotes the concatenated vertices of the facial surface and bone meshes, and its subscript indicates the corresponding space, \eg, $C$ indicates that it lies in the canonical space $\domain{C}$, and $M$ in the identity's material space $\domain{M}^{id}$.} We uniformly sample \review{$N_e$} points within $\domain{M}^{id}$ for regularization, and $\lambda_e$ is a weight.  Using \eqnref{eq:mapping_loss}, $\phi^{id}$ is pre-trained and fixed for each identity during actuation training.

To make the model generalize better, we train the canonical actuation function $\mathcal{A}(\cdot)$ to learn a tensor field that is agnostic to the geometry of each identity's face. The actual simulation uses a warped actuation field defined as:
\begin{equation}
    \label{eqn:warpActuation}
	\tilde{\mathcal{A}}^{id}(\x) = \R_{\phi^{id}}(\x)^{\top}\mathcal{A}(\phi^{id}(\x))\R_{\phi^{id}}(\x),
\end{equation}
where $\R_{\phi^{id}}$ is the rotational component of the Jacobian of $\phi^{id}$ at $\x$.
While a single canonical field $\mathcal{A}(\cdot)$ is trained for all identities, each one has their own, identity-adapted $\tilde{\mathcal{A}}^{id}(\cdot)$.
The intuition for this idea is that the learned canonical contractile direction should also be the same (aligned) for all identity within $\domain{C}$. For more details, please refer to the supplementary material.

\begin{figure*}
    \centering
    \includegraphics[width=1.\textwidth]{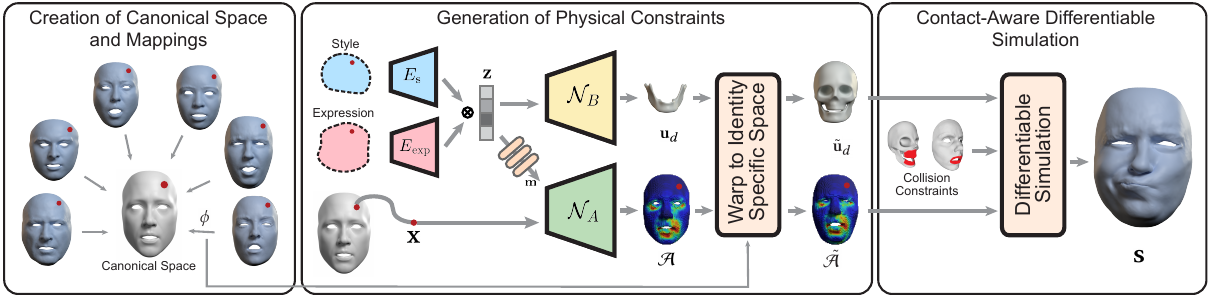}
    \caption{We illustrate our pipeline, consisting of the creation of a canonical space for training, our style- and expression-conditioned network for generating multi-identity physical constraints, and the contact-aware differentiable simulation.
    }
    \label{fig:pipeline}
\end{figure*}

\subsection{\review{Generation of Physical Constraints}}
Our method trains a unified implicit actuation network that defines the actuation tensor field solely in the shared canonical space $\domain{C}$, while also accommodating individual styles. Learning within $\domain{C}$ eases the burden to adapt to different material spaces. This streamlines its ability to discern common features across various identities and improves the transferability of the learned actuation mechanism.
To integrate individual nuances, we assign a distinct style code to each identity. Optimized alongside the network, this style code facilitates the utilization of all training data from all identities to train a singular network, thereby enhancing its generalization to unseen blendweight vectors originating from any actor's blendshape basis.

As shown in the pipeline,~\figref{fig:pipeline} (center), 
given the sampled expression code and the style code, 
two MLPs are used to map them to higher dimensional latent space, which are concatenated to get the activation code $\z$.
This code is used as the input to the implicit actuation network $\network_{A}$ to control the canonical actuation tensor field $\mathcal{A}$ through modulation. The modulation code $\mathbf{m}$ is generated from $\z$ by a tiny MLP network. $\mathcal{A}$ is then warped to the identity-specific material space $\domain{M}^{id}$ by the mapping function $\phi^{id}$, as $\tilde{\mathcal{A}}^{id}$.
At the same time, $\z$ is the direct input to another network $\network_{B}$ to get the jaw transformation from the rest state.
 \review{Analogous to actuation learning, we train $\network_{B}$ within the canonical jaw coordinate system. This allows it to be applied to any identity's jaw $\tilde{\mathbf{u}}^{id}_d$ in the identity's jaw coordinate system.}
Finally, the generated physical constraints are fed into the differentiable simulator to get the embedded facial geometry $\s$.  During training, we optimize for one style code per input identity, and use the per-frame blendweights as the expression code, together with the per-frame reconstructed geometry as the target of the simulation.  After training, the model can be driven with any expression and style code, for any of the seen identities. 
The objective for training our system consists of three terms.
The first term is to minimize the positional and normal difference between the captured and embedded facial geometry.
\begin{equation}
    \label{eq:loss1}
        \mathcal{L}_{\text{geo}} = ||\s - \s_{\text{captured}}|| + ||1 - \mathbf{n} \cdot \mathbf{n}_{\text{captured}}||.
\end{equation}
The second term is to regularize the actuation intensity to be as little actuated as possible,
\begin{equation}
    \label{eq:loss2}
        \mathcal{L}_{\text{act}} = \frac{1}{N_a}\sum_{i=1}^{N_a} ||\A_{i} - \mathbf{I}||,
\end{equation}
where $\mathbf{I}$ is the identity matrix, \review{and we sample $N_a$ actuation tensors}. This term can also stabilize the simulation and therefore the training process.
The third term is to enforce the smoothness by constraining the Lipschitz constant of the network to be small. 
Since the output of the network is a tensor, 
penalizing the gradient using auto differentiation is not trivial.
\citeN{liu2022learning} propose a more effective way by 
defining the smoothness energy solely based on the weights $\W$ of the network regardless of the inputs.
The loose Lipschitz bound $c$ of a $L$-layer fully-connected network
can be penalized directly in a differentiable manner by augmenting each layer 
with a Lipschitz weight normalization layer \cite{liu2022learning} that has a trainable Lipschitz bound $c_{i}$.
Therefore the third loss is defined as:
\begin{equation}
    \label{eq:loss3}
        \mathcal{L}_{\text{lip}} = \prod_{i=1}^{L} \text{softplus}(c_{i}),
\end{equation}
The overall loss is the weighted sum of the three terms:
\begin{equation}
    \label{eq:loss}
        \mathcal{L} = \mathcal{L}_{\text{geo}} + \lambda_{\text{act}} \mathcal{L}_{\text{act}} + \lambda_{\text{lip}} \mathcal{L}_{\text{lip}}.
\end{equation}
where $\lambda_{\text{act}}$ and $\lambda_{\text{lip}}$ are the weights for the two regularization terms.
For more details, please refer to the supplementary material.

\subsection{\review{Contact-Aware Differentiable Simulation}}
\label{subsec:collisions}
Our simulation framework consists of three energy terms: shape targeting, bone attachment, and contact energies. 
The first two terms are based on Projective Dynamics \cite{Bouaziz2014ProjectiveSimulation}, where the local constraint can be generally represented as follows:
\begin{equation}
\label{eq:local}
E_i(\u) = \min_{\y_i} \frac{\omega_i}{2} ||\mathbf{G}_i \mathbf{S}_i \u - \mathbf{B}_i \y_i||_F^2 \;\; \text{s.t.} \;\; C_i(\y_i) = 0,
\end{equation}
where $\u$ denotes the simulation vertices, $\omega_i$ is a weight coefficient, and $\y_i$ an auxiliary variable, embodying the target position.
$\mathbf{S}_i$ is a selection matrix choosing DOFs involved in $E_i$.
$\mathbf{G}_i$ and $\mathbf{B}_i$ are designed to facilitate the distance measure.
The total energy $E(\u)$ is the sum of all these local constraints.
This type of energy is solved using a local-global scheme. 
In the local step, we solve for all the $\y_i$ in parallel.
In the global step, we solve the following linear system derived by setting $\nabla E = 0$:
\begin{equation}
\label{eq:global}
\left(\sum_i \omega_i \mathbf{S}_i^{\top} \mathbf{G}_i^{\top} \mathbf{G}_i \mathbf{S}_i\right) \u=\sum_i \omega_i \mathbf{S}_i^{\top} \mathbf{G}_i^{\top} \mathbf{B}_i \y_i
\end{equation}

For the collision model, we adopt the IPC model \cite{li2020incremental} for its effectiveness, robustness, and versatility.
Without adding some inequality constraints explicitly,
the IPC model exploits a smoothly clamped barrier potential to penalize collision:
\begin{equation}
    B_{i}\left(d_i\right)= \begin{cases}-\left(d_i-\hat{d}\right)^2 \ln \left(\frac{d_i}{\hat{d}}\right), & 0<d_i<\hat{d} \\ 0, & d_i \geq \hat{d}\end{cases}
\end{equation}
where $d_i$ is the unsigned distance between the given pair $i$ of surface primitives, \ie, vertex-triangle or edge-edge, and
$\hat{d}$ is a user-defined tolerance of the collision resolution. 
As long as we ensure the collision-free state at the beginning of the simulation, 
using such potential forestalls any interpenetration afterwards.
Our idea is to use this barrier energy in our quasistatic facial simulation framework by solving the following optimization problem:
\begin{equation}
\label{eq:collision}
\min_{\u} E(\u) + B(\u),
\end{equation}
where $B(\u)$ is the sum of all the barrier potentials for all the collision pairs constructed from $\u$.
However, solving this optimization is not easy. $B(\u)$ is a highly nonlinear energy, and not PD-based.
For consistency, we project $B(\u)$ to its local hyperparaboloid centered at $\hat{\u}$, \ie, from the last iteration, using Taylor expansion:
\begin{equation}
\label{eq:hyperparaboloid}
\hat{B}(\u) = 
B(\hat{\u}) + 
\nabla B(\hat{\u})^{\top} (\u - \hat{\u}) + 
\frac{1}{2} (\u - \hat{\u})^{\top} \nabla^2 B(\hat{\u}) (\u - \hat{\u}).
\end{equation}
After projection, we add $\nabla \hat{B} = 0$ into the global step of \eqnref{eq:global}, giving the following new linear system:
\begin{equation}
\label{eq:global2}
    \Biggl(
    \nabla^{2} B + 
    \underbrace{\sum_i \omega_i \mathbf{S}_i^{\top} \mathbf{G}_i^{\top} \mathbf{G}_i \mathbf{S}_i}_{\mathbf{K}}
    \Biggr) \u =
    \nabla^2 B \hat{\u} - \nabla B + 
    \sum_i \omega_i \mathbf{S}_i^{\top} \mathbf{G}_i^{\top} \mathbf{B}_i \y_i,
\end{equation} 
where we remove the explicit evaluation of $\nabla B$ and $\nabla^2 B$ for simplicity.
We refer the readers to literature \cite{li2020incremental,yang2022implicit} for the details on the construction of the hessians.
In \eqnref{eq:global2}, $\mathbf{K}$ is a Laplacian-like positive definite symmetric matrix, that remains fixed throughout the simulation.
$\nabla^2 B$ , on the other hand, constantly changes but only affects a small portion of the left-hand side.
To maintain positive definiteness of the linear system, we project $\nabla^2 B$ to the cone of positive semidefinite matrices, as in Li et al.~\shortcite{li2020incremental}.
This linear system can then be efficiently solved using the preconditioned conjugate gradient method, 
with prefactorized $\mathbf{K}$ serving as the preconditioner. In addition, we use continuous collision detection to ensure that the system 
remains penetration-free at all iterations, and line search to guarantee convergence when necessary.
To improve the performance, we mark out the collision-prone regions (e.g. the lips) on the embedded facial surface and only consider those regions when constructing ${B}$, as shown in \figref{fig:pipeline}. 
\review{Since we use embedded simulation for its efficiency, the finer collision proxy $\mathbf{p}$ for constructing $B$ is linearly embedded in the coarser simulation mesh $\u$, \ie, $\mathbf{p} = \mathbf{W}\u$
. Consequently, we get $\nabla^2 B(\u) = \mathbf{W}^{\top} \nabla^2 B(\mathbf{p}) \mathbf{W}$. The linear mapping also guarantees the linear trajectories of $\mathbf{p}$, making it possible to use CCD in our optimization framework.}

\subsection{Implementation Details}
\label{subsec:implementation}

For the simulator during training, we detach our collision model for speed-up, and use Projective Dynamics~\cite{Bouaziz2014ProjectiveSimulation} as our forward solver, 
and differentiable Projective Dynamics~\cite{Du2022DiffPD:Dynamics} as backward solver, where we adopt similar optimization strategies as in Yang et al.~\shortcite{yang2022implicit}.
During inference, we use our proposed solver with the collision model attached.
For the actuation network, we use a similar architecture as Yang et al.~\shortcite{yang2022implicit} but with two main differences.
First, we replace all the SIREN layers except the first with GeLU layers, since we found SIREN layers to produce unstable and noisy results that were extremely sensitive to initialization. Second we apply a tanh activation function for the modulating code $\mathbf{m}$ that directly modulates the actuation network backbone, and can be used for T-SNE visualization, as shown in \figref{fig:lipschitz_regularization}. \review{For the visualization of actuation values, we use the Frobenius distance between the actuation matrix and the identity matrix.}
For all experiments shown in \secref{sec:experiments}, we use PyTorch~\cite{PyTorch2019} on the NVIDIA RTX 2080Ti with the simulator integrated as a layer. For more details, please refer to the supplementary material.

\section{Experiments}
\label{sec:experiments}

We start by evaluating our model learned on multiple identities, and then demonstrate several applications like style transfer, animation retargeting and other physical effects.

\begin{table}
\small
    \caption{Numerical evaluation of different models on the test set.}
\vspace{-3mm}
    \begin{tabular}{l|c}
        \toprule
        \textbf{Methods}&                       \textbf{Vertex Error (mm) $\downarrow$}           \\ \hline
        \review{Linear Blendshape Rig}                   & \review{0.5549}                              \\  
        Model-N   (w.o. CS)	                    &  0.4119                              \\ 
        Model-CS    (w. CS)	                    &  0.4105                             \\
        Model-CSL   (w. CS + Lip)	            &  0.4055                             \\
        Model-CSW   (w. CS + Warp)              &  0.4020                             \\
        Model-CSWL  (w. CS + Warp + Lip) \review{[Ours]}       &  \textbf{0.3849}                    \\
        \review{Model-CSWL  (Displacement Only)}         & \review{0.4313}                              \\  
        Model-CSWL  (Single-Identity)           & 0.4218                              \\
        \review{\cite{yang2022implicit}}                 & \review{0.4143}                               \\    
        \bottomrule
    \end{tabular}
    \label{tab:numerical_evaluation}
\vspace{-5mm}
\end{table}

\subsection{Model Evaluation}
\label{subsec:modelevaluation}
\paragraph{Multi-identity Model.}
In this section, we quantitatively and qualitatively evaluate our multi-identity model.  For quantitative evaluation we use a reconstruction error metric, which is the average Euclidean distance between the ground truth and the reconstructed mesh vertices. As outlined in Section \ref{subsec:trainingData}, three sequences, one for each of the three identities from the training set, are allocated for this purpose. However, it is important to note that the reconstruction accuracy only reflects the performance of the model on within-identity animation tasks, where ground-truth performance data is available.
Cross-identity animation tasks including retargeting and style transfer %
provide a more comprehensive test of the model's generalization capabilities, for which we provide qualitative comparisons.

There are three main components in our model that we can evaluate,
including the canonical space, the warp operation,
and the Lipschitz regularization. 
First, we evaluate our canonical space and the warp operation.
The Canonical space is expected to let the model learn a more unified muscle structure, and thus ease the burden to adapt to different material spaces.
The warp operation is further introduced for better generalization.
For evaluation, we train three models with different simplifications. 
The first (Model-N) is a basic multi-identity model trained on separate material spaces, without the canonical space. The second (Model-CS) uses the canonical space for multi-identity learning, but without a per-identity warp function. Lastly, Model-CSW employs the canonical space and warp operation for multi-identity learning.  \tabref{tab:numerical_evaluation} shows the corresponding errors when reconstructing the test set. Here, Model-CSW exhibits the best performance of these 3 models, and Model-N the worst. 

In order to show the benefits of our Lipschitz regularization,
we further train two more models, Model-CSL and Model-CSWL that are equipped with the Lipschitz regularization on top of Model-CS,
and Model-CSW. As shown in \tabref{tab:numerical_evaluation}, the effect of the warp operation is consistent while the Lipschitz regularization makes the model generalize to the test set even better.
Our proposed full model (Model-CSWL) achieves the lowest error on the test set.

\begin{figure}[b]%
    \centering
    \includegraphics[width=\linewidth]{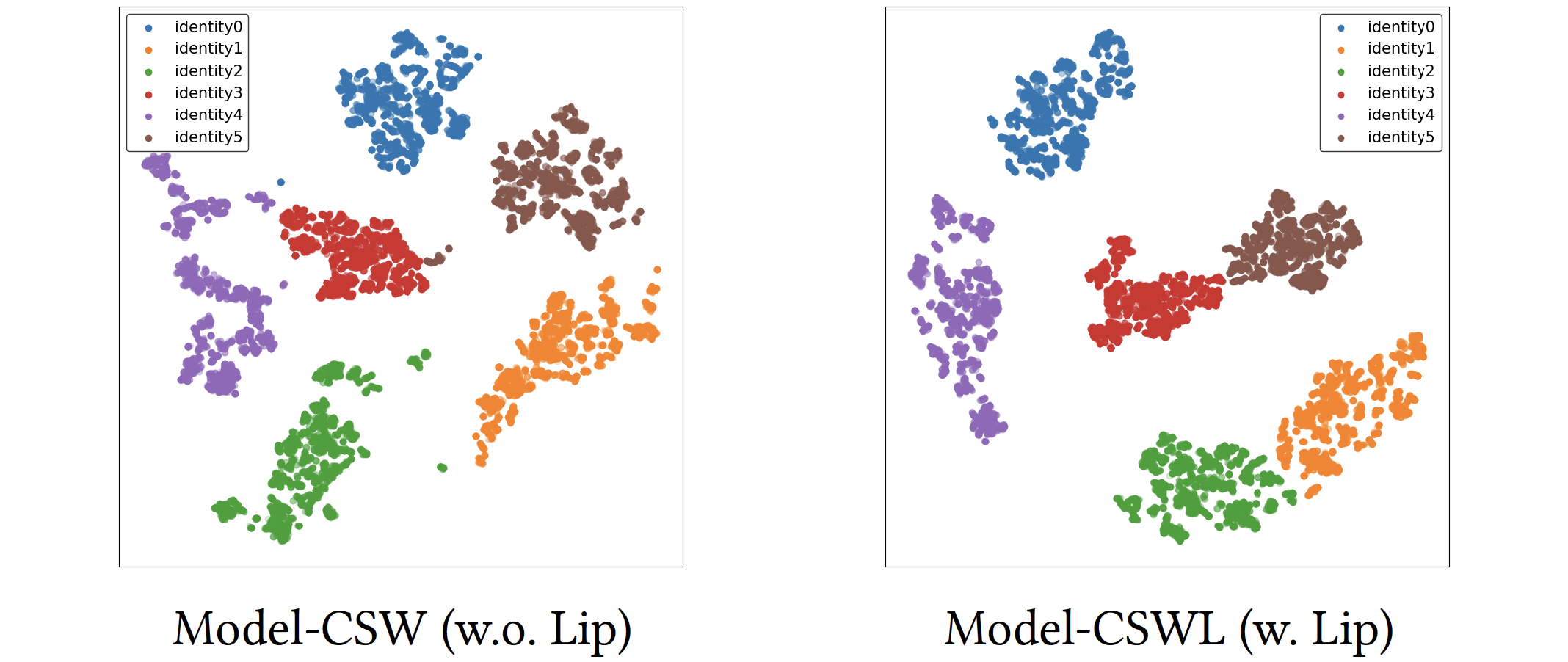}
    \caption{Our model can disentangle identity style vs. expression, as shown by a T-SNE plot of actuation modulation codes $\mathbf{m}$.  With Lipschitz regularization (right) achieves better disentanglement than without (left).}
    \label{fig:lipschitz_regularization}
\end{figure}

To further evaluate the Lipschitz regularization, we analyze our model's ability to disentangle expression and style with and without regularization.  To this end, we employ T-SNE to visualize a comprehensive set of actuation modulating codes $\mathbf{m}$, generated by pairing the expression blendweight vectors of the entire training dataset (across all identities) with each learned style code. Successful disentanglement would mean that these modulation codes should cluster by identity style, whether the expression blendweights come from the particular identity or another identity. As demonstrated in \figref{fig:lipschitz_regularization}, Model-CSWL exhibits superior style and expression disentanglement compared to Model-CSW, where points drifting outside clusters or misallocated to incorrect clusters can be observed.

\begin{figure}[t]
    \centering
    \includegraphics[width=1.0\linewidth]{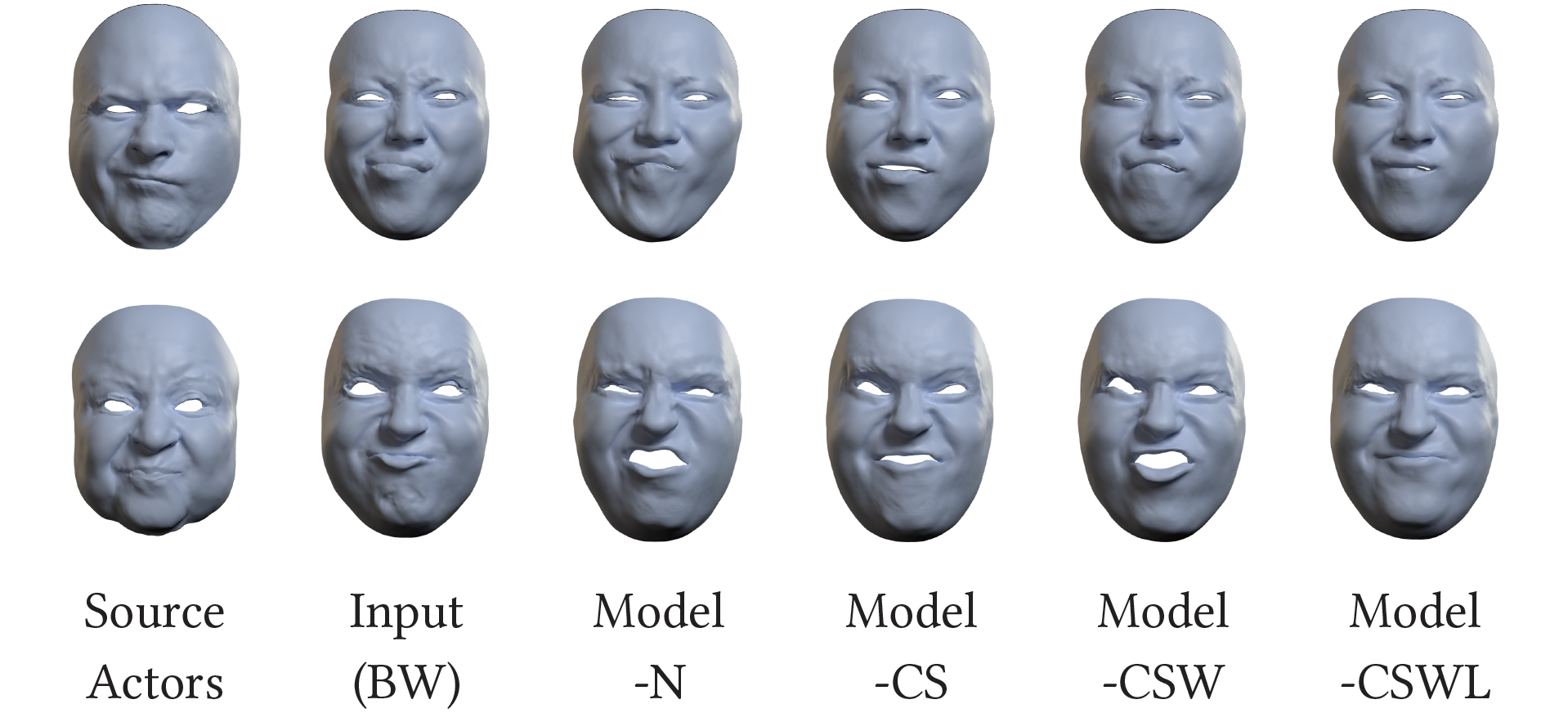}
    \caption{Qualitative evaluation of our model components on two retargeting examples. The input is the unseen blendweight (BW) vector of a source actor (col 1), illustrated with simple blending on the target identity's blendshape model (col 2).  Among the model variants, our proposed model (Model-CSWL) achieves the most natural results with fewer artifacts.}
    \label{fig:ablation}
\end{figure}

Qualitative comparisons of these models, in terms of cross-identity animation quality, are depicted in \figref{fig:ablation}. 
The full model, Model-CSWL, generates more natural results, while the other methods exhibit more expression mismatches and artifacts.

\begin{figure}[b] %
    \centering
    \includegraphics[width=1.0\linewidth]{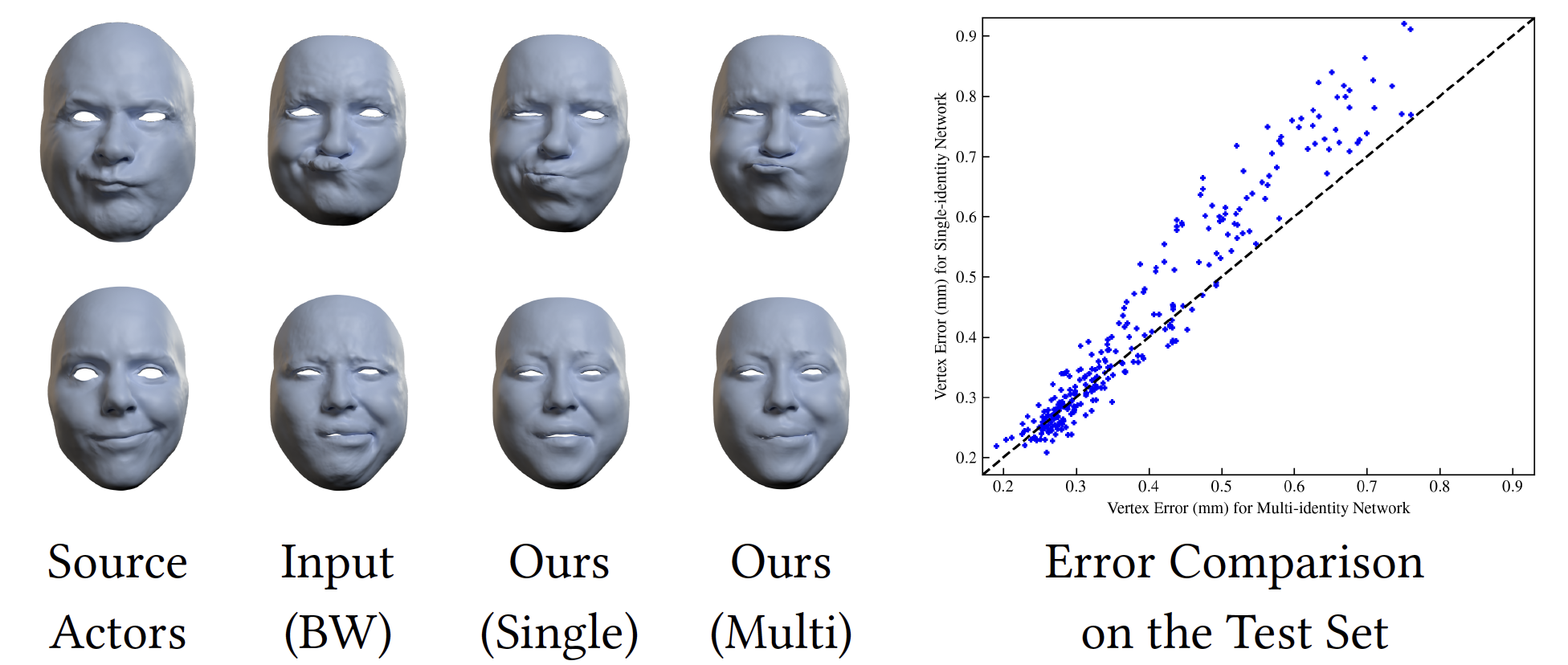}
    \caption{Comparison between single-identity and multi-identity models on a retargeting example (left), and visualization of reconstruction errors (right).  Our multi-identity model achieves lower errors and more natural shapes.}
    \label{fig:single_vs_multiple}
\end{figure}

\paragraph{Comparison with Single-identity Model.}
We also compare our model to our single-identity variant, both quantitatively and qualitatively.
To ensure a fair comparison, we train three single-identity models—each for a test identity—with the same specifications as our multi-identity model. The key distinction is the data source: single models use data from one identity, while the multi-identity model uses data from several. As indicated in \tabref{tab:numerical_evaluation}, our multi-identity model has a lower error on the test set. The visualization of the error points, collected from the averaged vertex error per frame, is shown on the right half of the \figref{fig:single_vs_multiple}.
Note that the error points are biased towards the upper part of the diagonal dotted line. We also compare cross-identity animation results. As shown on the left half of \figref{fig:single_vs_multiple}, our multi-identity model achieves better shape matching and more natural results. \review{These experiments show that when equipped with more data from other identities, our model generalizes better and is more effective, especially when it comes to cross-identity animation.}

\begin{figure}[t] %
    \centering
    \includegraphics[width=1.0\linewidth]{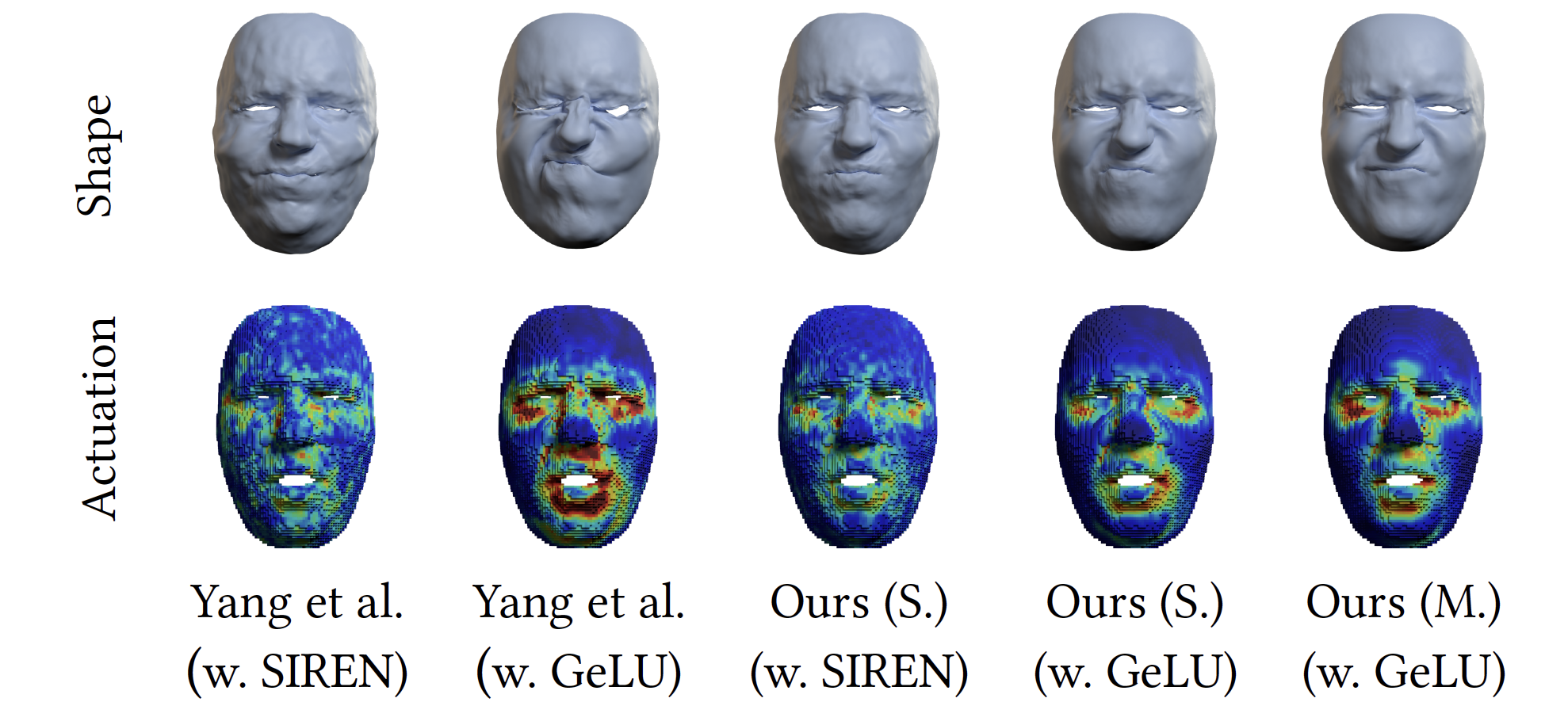}
    \caption{We compare our model trained on a single identity (S.) and multi-identities (M.) to the model of Yang et al.~\shortcite{yang2022implicit}, with both SIREN and GeLU activations.  Our multi-identity model with GeLU exhibits less artifacts.}
    \label{fig:compare_with_yang}
\end{figure}
It is worth noting that our single-identity model is adapted from the multi-identity model, and thus has the two main differences from Yang et al.~\shortcite{yang2022implicit} that we mentioned in \secref{subsec:implementation}.  Namely, it replaces all but the first SIREN layers with GeLU layers and adds a tanh activation after modulation code.  \review{As shown in \figref{fig:compare_with_yang}, these changes largely alleviate artifacts visible in the method of Yang et al.~\shortcite{yang2022implicit}, while achieving comparable accuracy (see  \tabref{tab:numerical_evaluation})}.

\paragraph{Comparison with Other Models.}
\review{As reported in \tabref{tab:numerical_evaluation}, our model performs better than a linear blendshape rig that uses the same blendshape basis as ours. It also outperforms our displacement variant, which bypasses physics completely and directly learns the offset field. More importantly, our model supports style transfer and physical effects.}

\begin{figure}[t]%
    \centering
    \includegraphics[width=\linewidth]{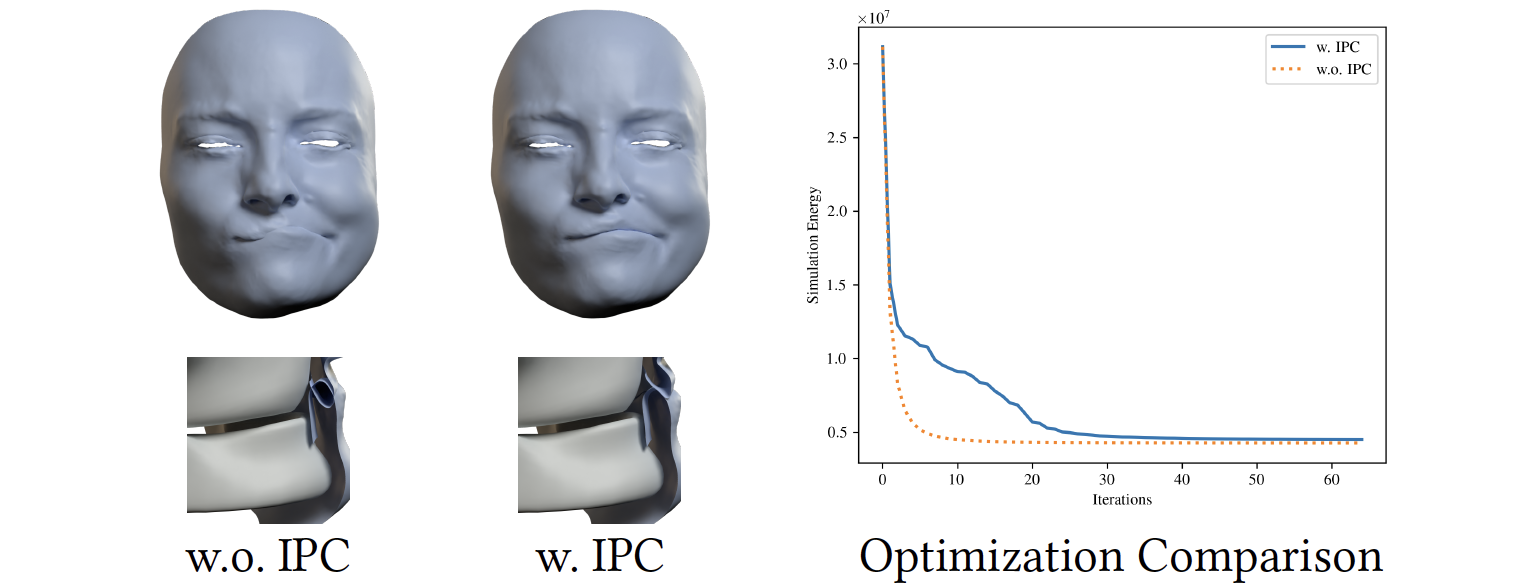}
    \caption{We juxtapose our method, with and without IPC. The left side illustrates a visual comparison, with the mouth cut-away displayed in the bottom row. The right side depicts the optimization curves.}
    \label{fig:collision_evaluation}
\end{figure}

\paragraph{Collision Model.}
The IPC model plays a critical role in helping achieve the \emph{collision-free} and \emph{converged} results, as shown in \figref{fig:collision_evaluation}.
Note that previous works commonly use the level-set method for consistency with the framework of Projective Dynamics.
However, as shown in Li et al.~\shortcite{li2020incremental}, this method is not robust and prone to penetration. 

In addition, we find that the IPC model can support various applications.
As shown in \figref{fig:differentiable_collision}, the IPC model's natural differentiability could be used to correct data.
Second, it naturally supports the simulation of friction effects, as shown in \figref{fig:friction}.

\subsection{Style Transfer}
\begin{figure}[t] %
    \centering
    \includegraphics[width=1.0\linewidth]{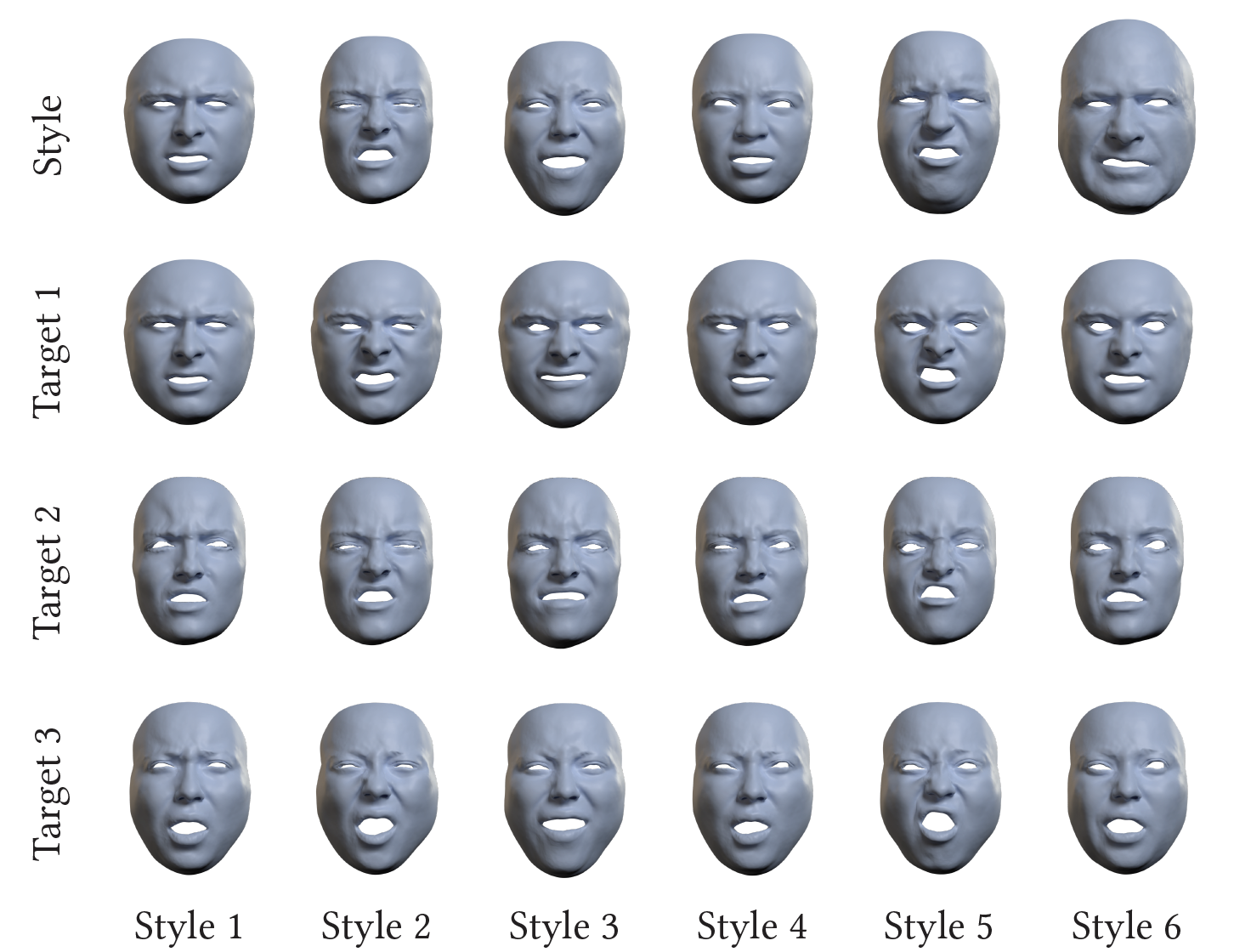}
    \caption{Here we demonstrate the application of Style Transfer.  We can transfer the expression style from any source identity to any target identity, while keeping the expression code fixed.}
    \label{fig:styletransfer}
\end{figure}

\begin{figure}[t]%
    \centering
    \includegraphics[width=1.0\linewidth]{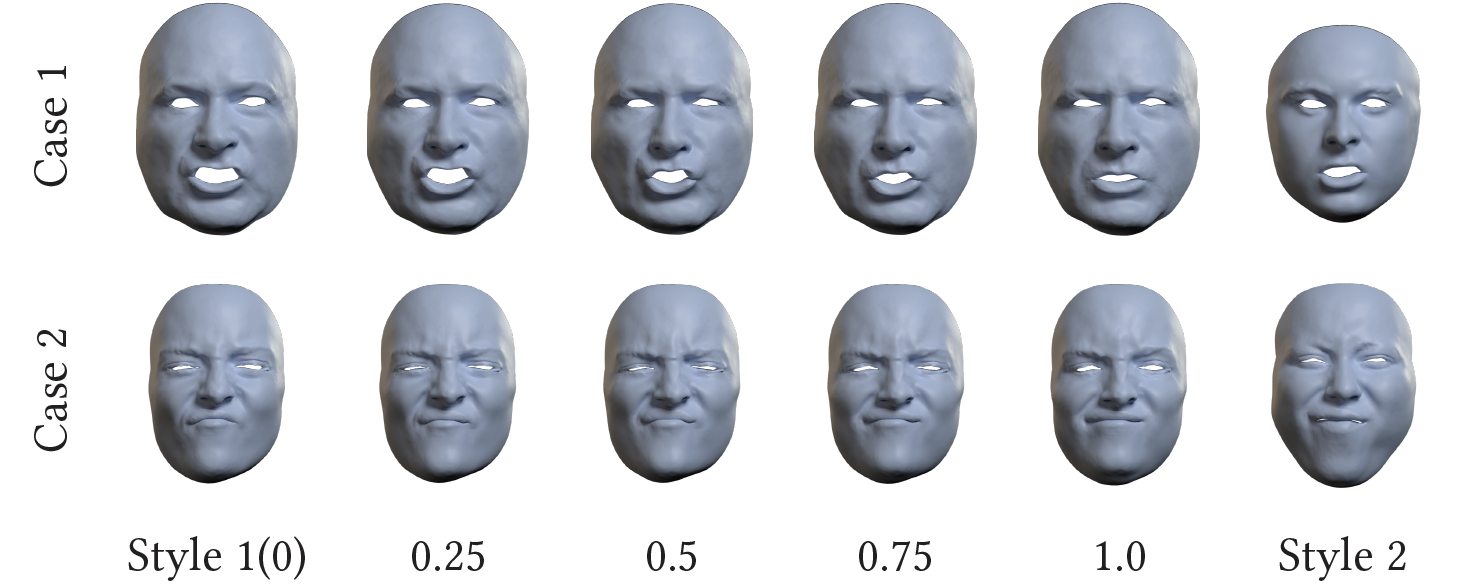}
    \caption{We can smoothly interpolate the style code from one style to another, demonstrated here for two examples where the style is slowly changed to that of identity 2, visualized on the geometry of identity 1.}
    \label{fig:styleinterpolation}
\end{figure}

\begin{figure}[t]%
    \centering
    \includegraphics[width=1.0\linewidth]{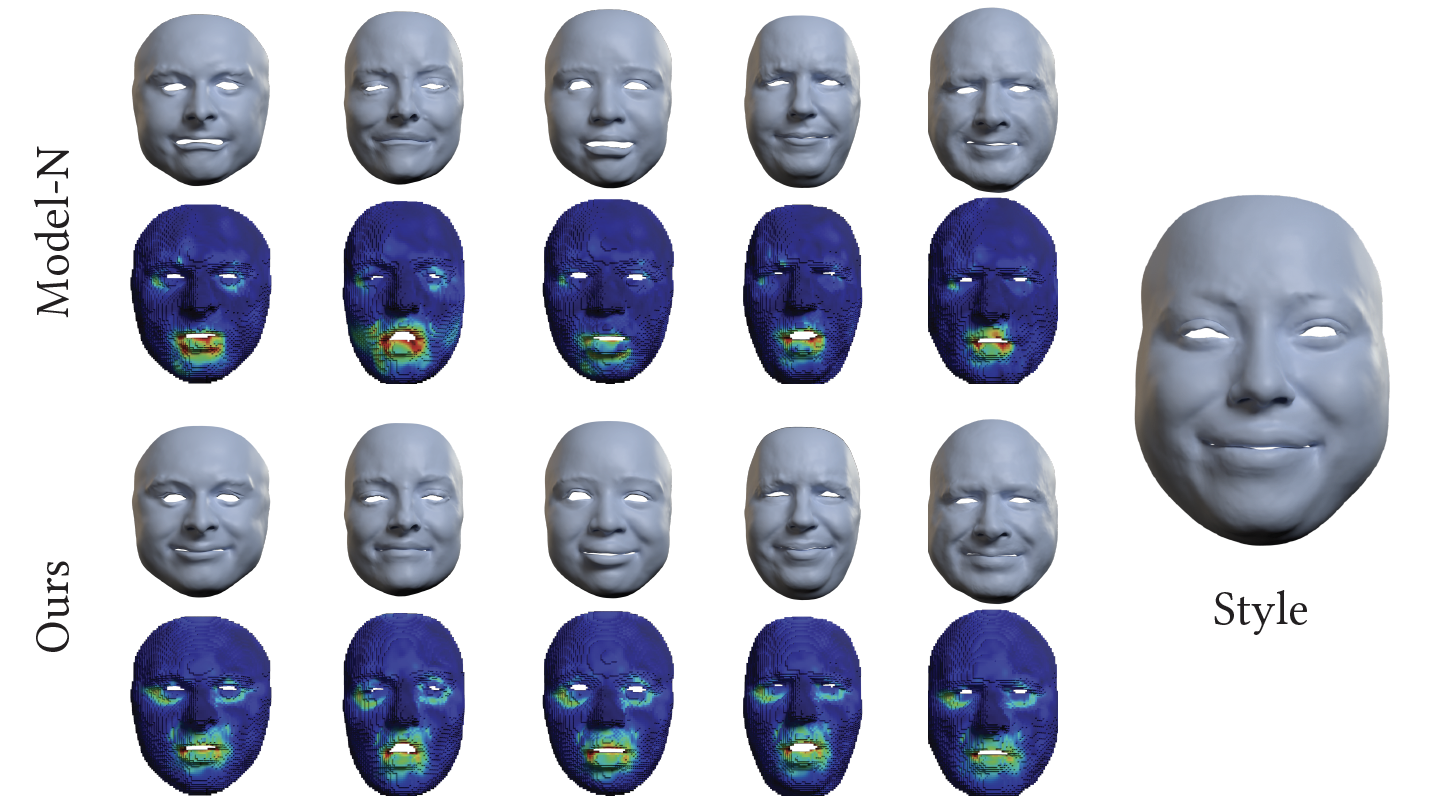}
    \caption{We show a style transfer comparison between our model and the model without canonical space (Model-N) for 5 identities. The actuation field is inconsistent across identities in Model-N for the same style.}
    \label{fig:styletransfercomparison}
\end{figure}

As demonstrated in \figref{fig:styletransfer}, our approach enables style transfer between different identities by utilizing the style code of a source identity with the geometry of a target identity. Notably, our method also facilitates style interpolation, as shown in \figref{fig:styleinterpolation}, a capability made possible by the Lipschitz regularization technique \cite{liu2022learning}. It's worth mentioning that our shared canonical space, $\domain{C}$, plays an essential role in this capability. \figref{fig:styletransfercomparison} offers a comparison between our method and a model devoid of the canonical space (Model-N). The results indicate that our approach ensures consistent actuation across different identities, while Model-N fails to deliver similar consistency and produces unnatural and inconsistent results. This is because, for the same style code, different identities in Model-N query the actuation tensors within their respective material spaces, leading to inconsistent actuation results. In contrast, our model successfully maintains actuation uniformity due to its learning within the shared canonical space, which supports transferability across different identities.

\subsection{Retargeting}
\begin{figure}[t]
    \centering
    \includegraphics[width=1.0\linewidth]{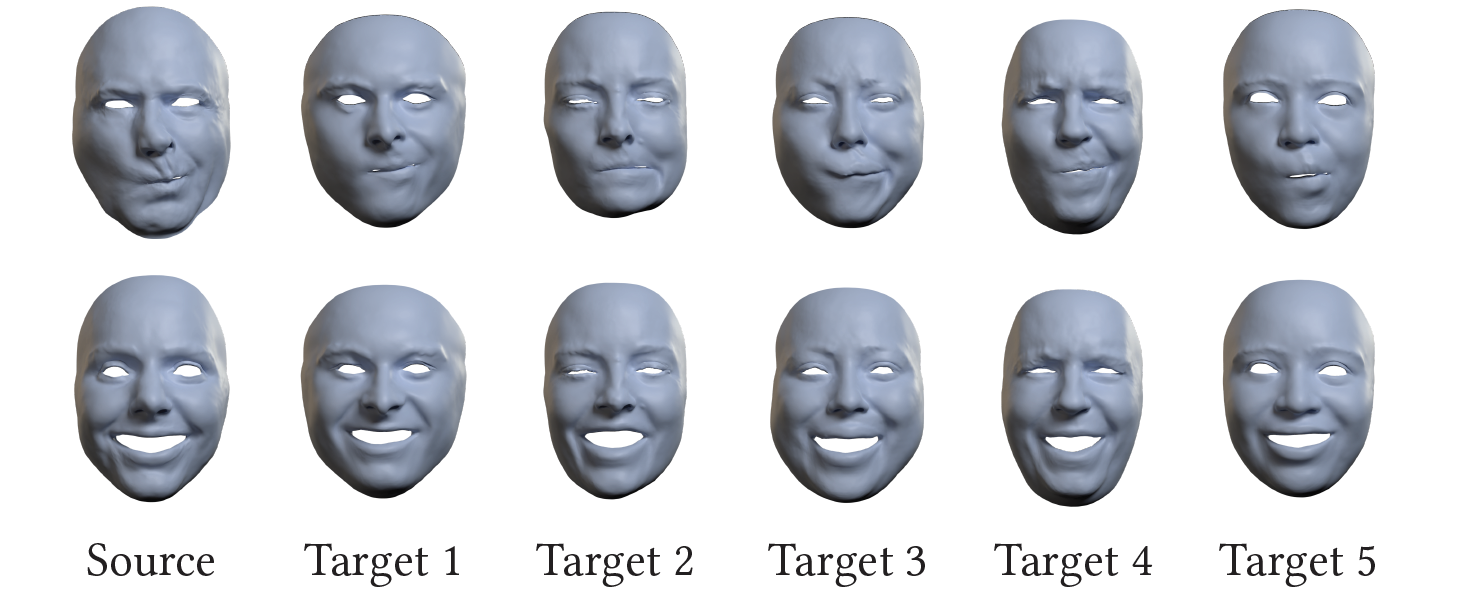}
    \caption{Here we show animation retargeting for 2 expressions from different source actors retargeted to 5 different target identities (in their own style).}
    \label{fig:retargeting}
\end{figure}

\begin{figure}[!htb]
    \centering
    \includegraphics[width=\linewidth]{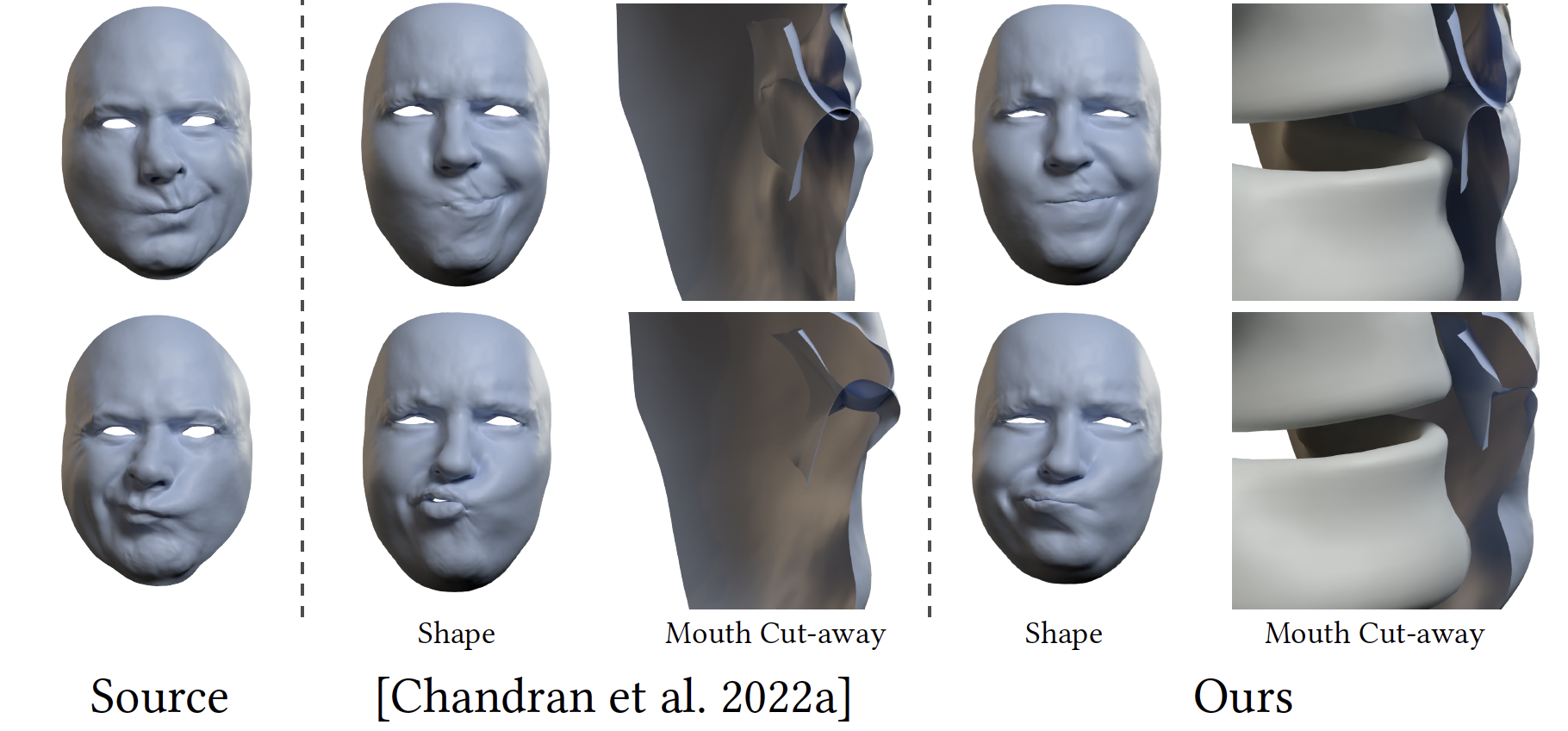}
    \caption{Comparing to the retargeting method of Chandran et al.~\shortcite{Chandran2022Karacast}, our method naturally resolves lip intersections in the retarget result.}
    \label{fig:comparekaracast}
\end{figure}

Our method can be used for facial animation retargeting thanks to the enhanced generalization from multi-identity learning, as shown in \figref{fig:retargeting}.  Retargeting is achieved by using the expression blendweights of one identity but the style and geometry of another identity.
Note that the lip penetration problems that are common in conventional methods~\cite{Chandran2022Karacast} can be naturally solved with our method, as shown in \figref{fig:comparekaracast}. %
In addition, our model can allow other physical effects on top of the retargeting results, as shown in \secref{sec:various_physical_properties}.

\subsection{Various Physical Effects}
\label{sec:various_physical_properties}
\begin{figure}[t]%
    \centering
    \includegraphics[width=1.0\linewidth]{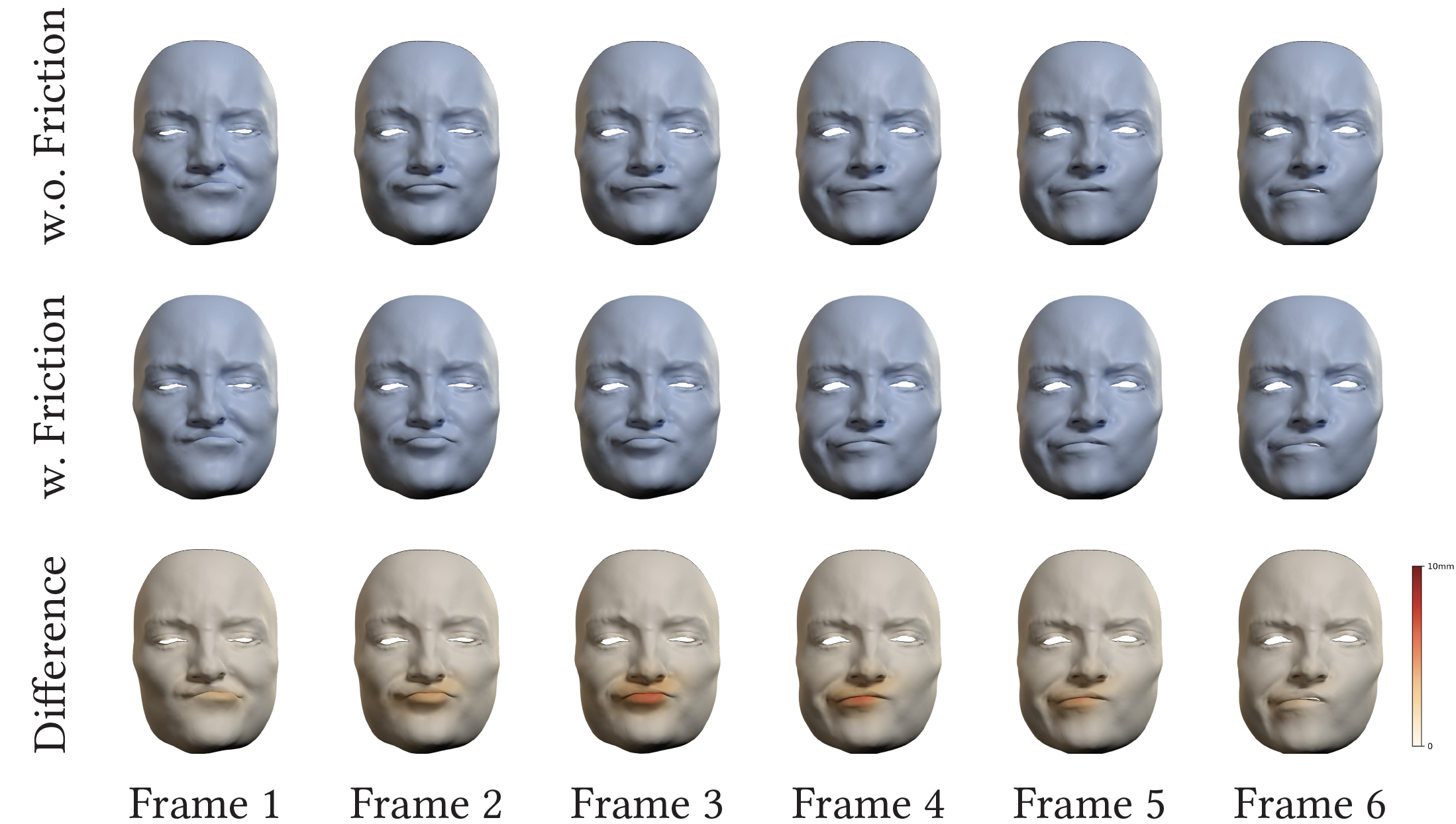}
    \caption{Our IPC contact model allows to simulate friction effects, shown here for the mouth region.}
    \label{fig:friction}
\end{figure}

\paragraph{Friction.}
Our model effectively simulates friction effects in the quasistatic setting. To enable this, for each frame, we initialize the simulation mesh and compute the \emph{lagged} sliding basis from the previous frame. This basis aids in formulating the smoothed static friction energy for the current frame, as in \cite{li2020incremental}. Our solver remains compatible, the only difference is to add the hessian and gradient from the friction energy into the optimization loop.
As shown in \figref{fig:friction}, the lower lip first slips inside and then pops out when no friction is applied, 
whereas it consistently moves along with the upper lip when friction is imposed upon it.

\begin{figure}[t]%
    \centering
    \includegraphics[width=1.0\linewidth]{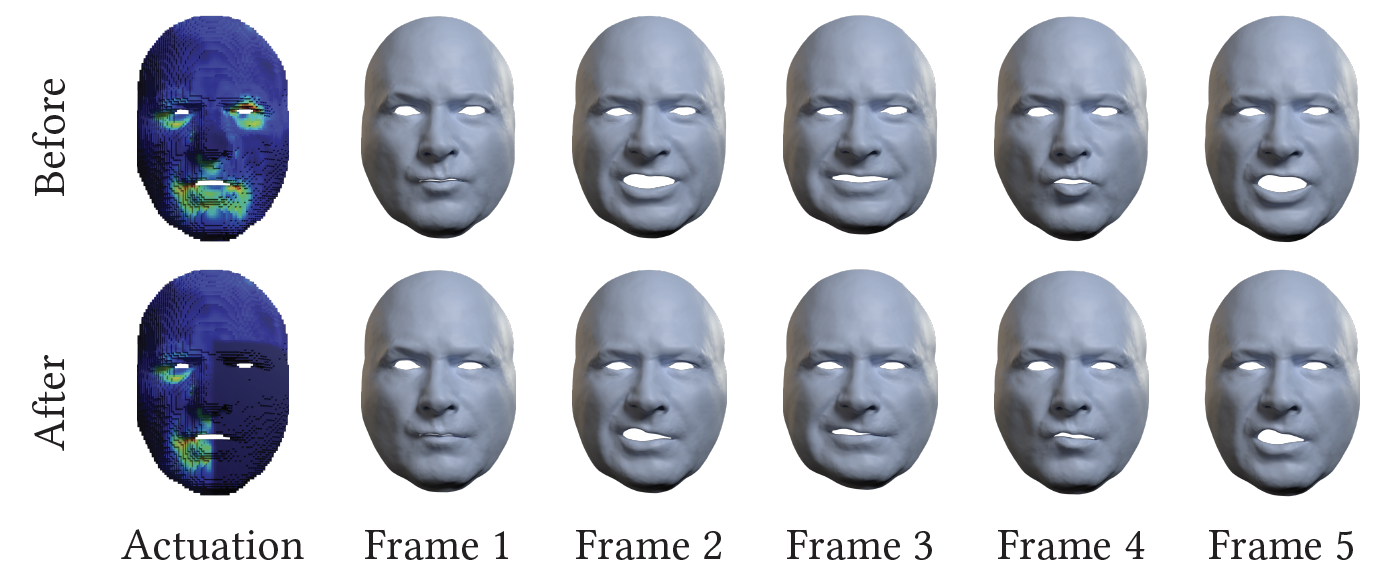}
    \caption{Our model can simulate artistic effects like facial paralysis.}
    \label{fig:paralysis}
\end{figure}

\paragraph{Paralysis.}
Our method also allows to add artistic effects like partial facial paralysis, as shown in \figref{fig:paralysis}.
Note that the paralyzed part is not completely fixed, but naturally deforms along with the other parts, \eg, the mandible movement,
which is in contrast with conventional methods, where the paralyzed part is more or less fixed, as in Chandran et al.~\shortcite{Chandran2022Karacast}.

\begin{figure}[t]
    \centering
    \includegraphics[width=1.0\linewidth]{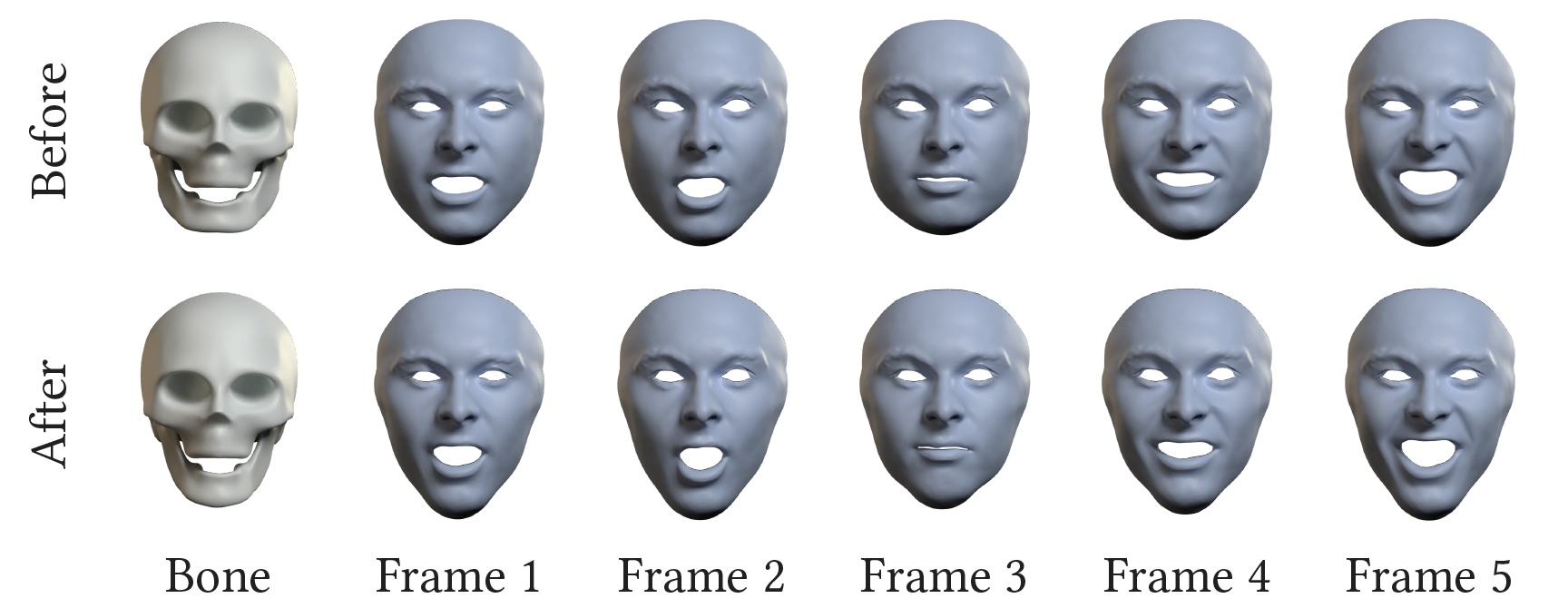}
    \caption{A physical effect that our model can achieve is osteotomy. Here we shrink the mandible bone and the facial animation naturally matches.}
    \label{fig:osteotomy}
\end{figure}

\paragraph{Osteotomy.}
Our model could also be used to simulate osteotomy, where we change the shape of the mandible and replay the actuation.
As shown in \figref{fig:osteotomy}, our model is able to achieve natural and realistic animation results.

\begin{figure}[t] %
    \centering
    \includegraphics[width=1.0\linewidth]{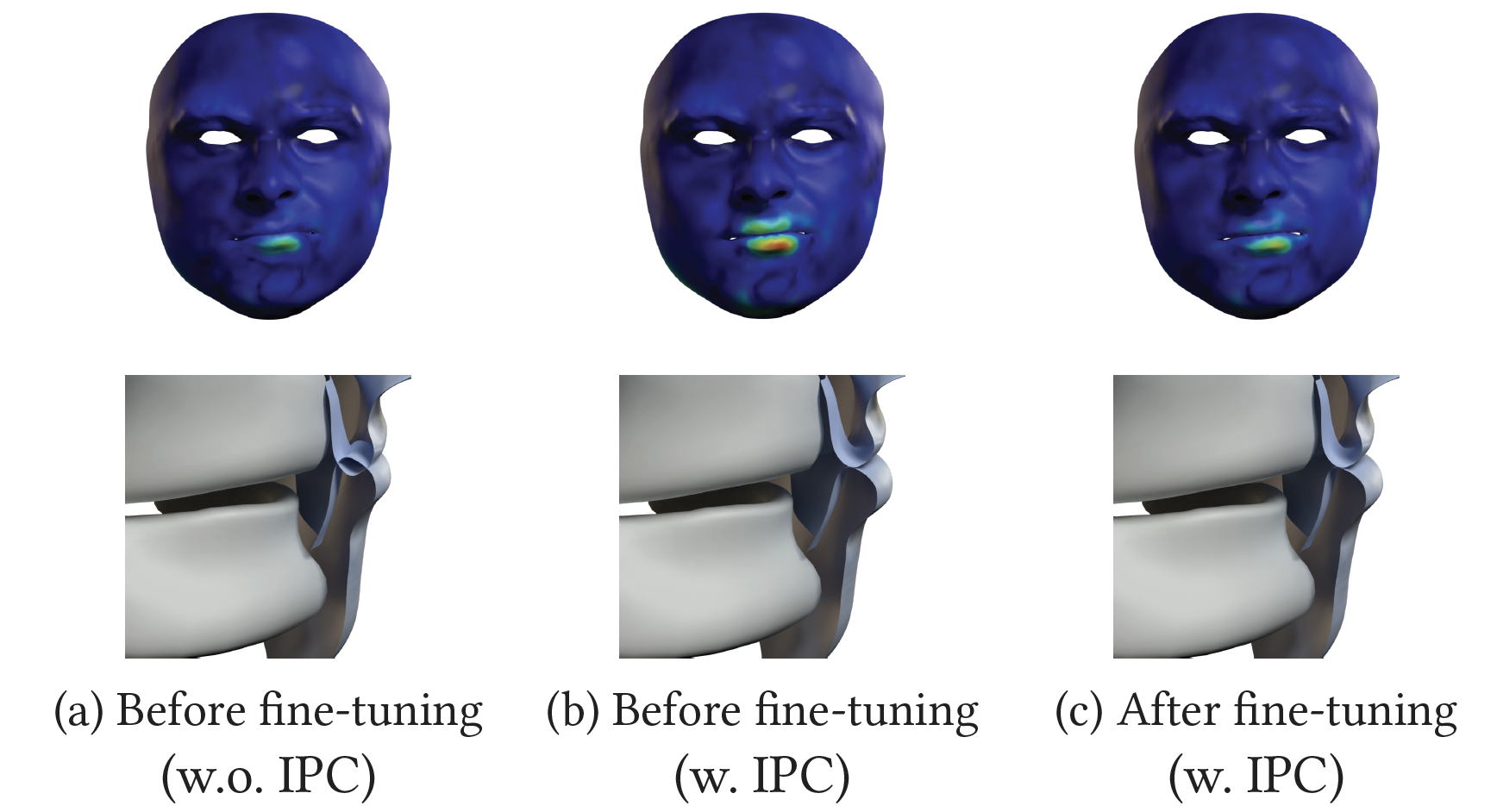}
    \caption{Our collision model supports differentiable collision handling, 
    allowing us to fine-tune the network for better fitting to captured data.%
    }
    \label{fig:differentiable_collision}
\end{figure}

\paragraph{Differentiable Collision.}
Our collision model naturally supports differentiability.
Since the captured data is prone to lip penetration, this can affect training results, as shown in \figref{fig:differentiable_collision} (a).
When we turn on the collision model during testing, the penetration is corrected but increases the error, as shown in \figref{fig:differentiable_collision} (b).
After we fine-tune the network with the collision model integrated, the error is reduced, as shown in \figref{fig:differentiable_collision} (c).

\section{Conclusion}
\label{sec:conclusion}

We propose a data-driven physical model that leverages multi-identity learning and can transfer performance and style across subjects. Our framework generalizes from animation examples seen on any of the trained identities to create an even richer and more expressive physics-based model for all subjects, and allows for manipulation of expression and style. 
We inherit some of the design limitations of prior work on data-driven face models  \cite{srinivasan2021learning, yang2022implicit}, such as being limited to the material constitutive model of shape targeting \cite{klar2020Shapetargeting} as opposed to more biomechanically accurate models. Due to hardware constraints and the computational demands of simulation, we use a small finite set of identities in our training set, thus we do not claim to span the space of human face identities and styles as exhaustively as we cover the space of expressions. That said, by virtue of the Lipschitz regularization, we retain some ability to interpolate between the identities in the training set. Finally, we have not aggressively optimized our pipeline in terms of efficiency of training, focusing rather on quality and generalization than speed of training.

\begin{acks}

    We thank the anonymous reviewers for their constructive comments.
    The work is supported by the 
    {Swiss National Science Foundation} under Grant No.: {200021\_197136}.
\end{acks}

\bibliographystyle{ACM-Reference-Format}
\bibliography{reference}

\clearpage
\section{Appendix}
This appendix contains more details on our implementation, additional results and comparisons with other methods.
\subsection{Mapping Function}

Given the the canonical space $\domain{C}$, the material space $\domain{M}$ of a target identity,
and a bijective mapping function $\phi$ between these two spaces:
$\phi: \x \in \domain{M} \rightarrow \X \in \domain{C}$,
and 
$\phi^{-1}: \X \in \domain{C} \rightarrow \x \in \domain{M}$. 
We define the actuation tensor field on $\domain{C}$ as $\mathcal{A}$, 
which will be warped to $\domain{M}$ as $\tilde{\mathcal{A}}$ to deform the target face.

The energy function defined on $\domain{C}$ is 
\begin{equation}
    E = \int_{\domain{C}} \frac{1}{2}\left\|\mathbf{F(\X)}-\mathbf{R^{*}(\X) \mathcal{A}(\X)}\right\|_{F}^{2} dV
\end{equation}
What if we directly push forward it to $\domain{M}$.
\begin{equation}
    \label{eqn:pushed_forward}
    E = \int_{\domain{M}} \frac{1}{2}\left\|
        \mathbf{F(\x)}\frac{\partial{\phi^{-1}(\X)}}{\partial \X} 
        -
        \mathbf{R^{*}(\x)} \mathcal{A}(\phi(\x))\right\|_{F}^{2} \left|\frac{\partial{\phi(\x)}}{\partial \x}\right|dv,
\end{equation}
where $\F(\x)$ is the deformation gradient measured in $\domain{M}$,
This is the typical way to view the simulation in different spaces (material space and Eulerian space).
However, in current setting, the canonical space $\domain{C}$ and the target space $\domain{M}$ 
are two independent spaces, this pushing forward is not meaningful, 
e.g., the deformation gradient should not be accumulated as $\mathbf{F(\x)}\frac{\partial{\phi^{-1}(\X)}}{\partial \X}$.
To see this, we can assume the actuation tensors are all identity matrices (no induced deformation).
Then, ideally, the target identity should remain undeformed. But if we use the energy function \eqref{eqn:pushed_forward} to run the simulator for the target identity, 
it will be dragged into the canonical space, which is not what we want. 

\begin{figure}[!t]
    \centering
    \begin{tabular}{
    P{0.3\linewidth}@{\hspace{3pt}}
    P{0.3\linewidth}@{\hspace{3pt}}
    P{0.3\linewidth}@{\hspace{3pt}}}
    \includegraphics[trim=0 0 100 450, clip, width=1\linewidth]{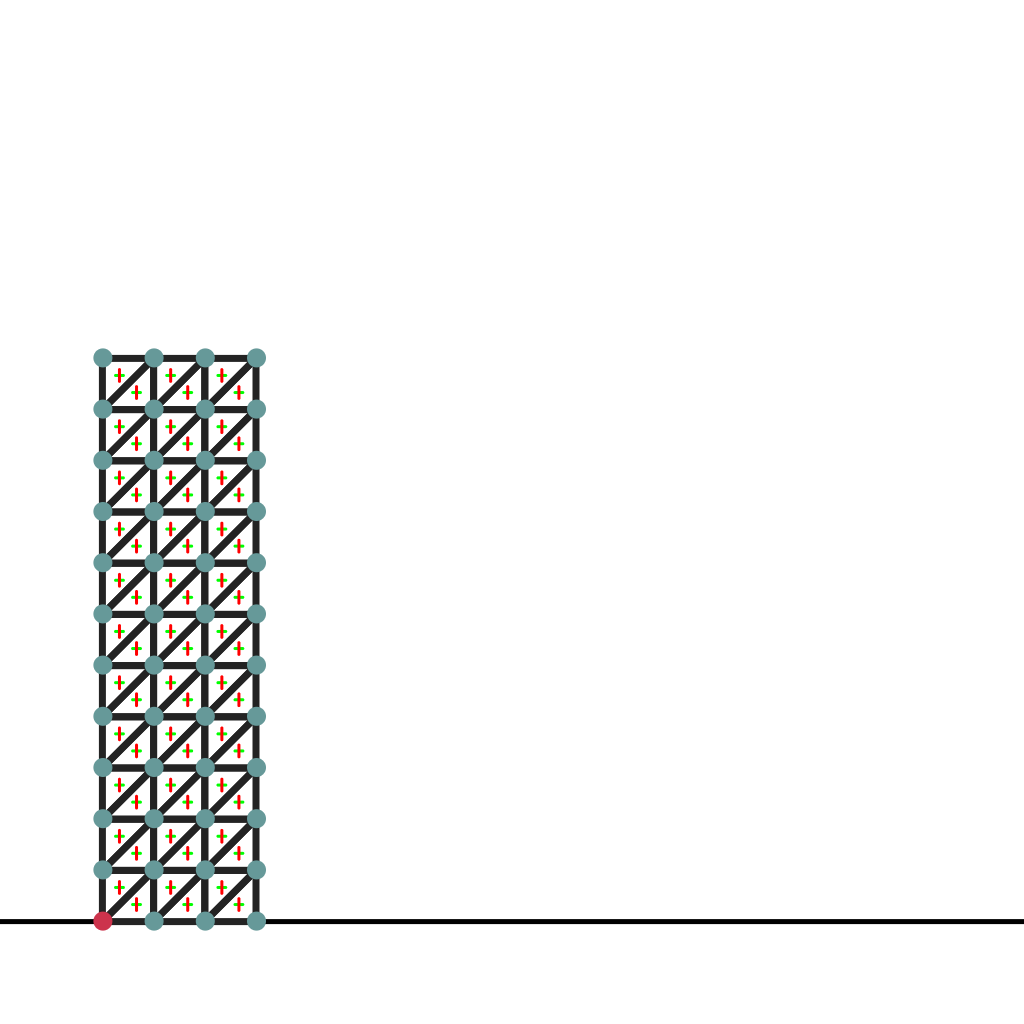}       &
    \includegraphics[trim=0 0 100 450, clip, width=1\linewidth]{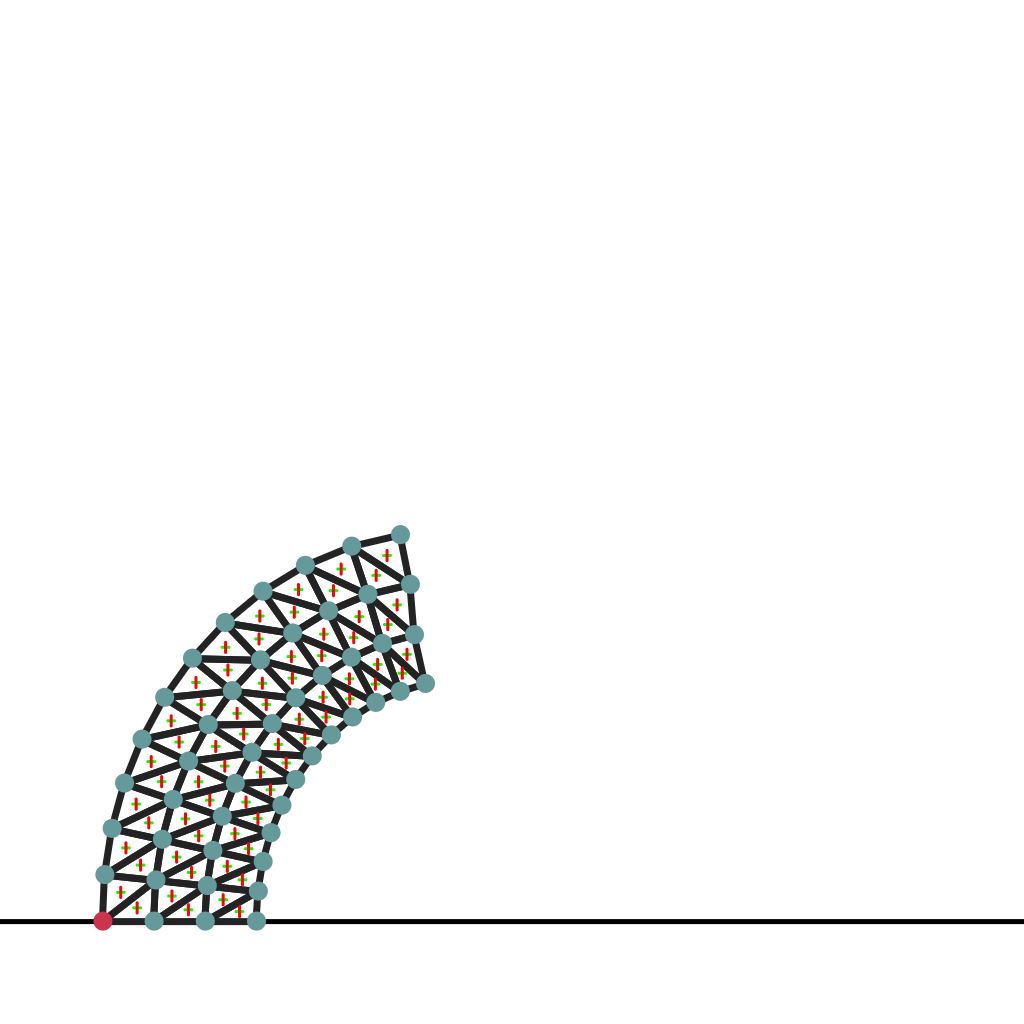}    &
    \includegraphics[trim=0 0 100 450, clip, width=1\linewidth]{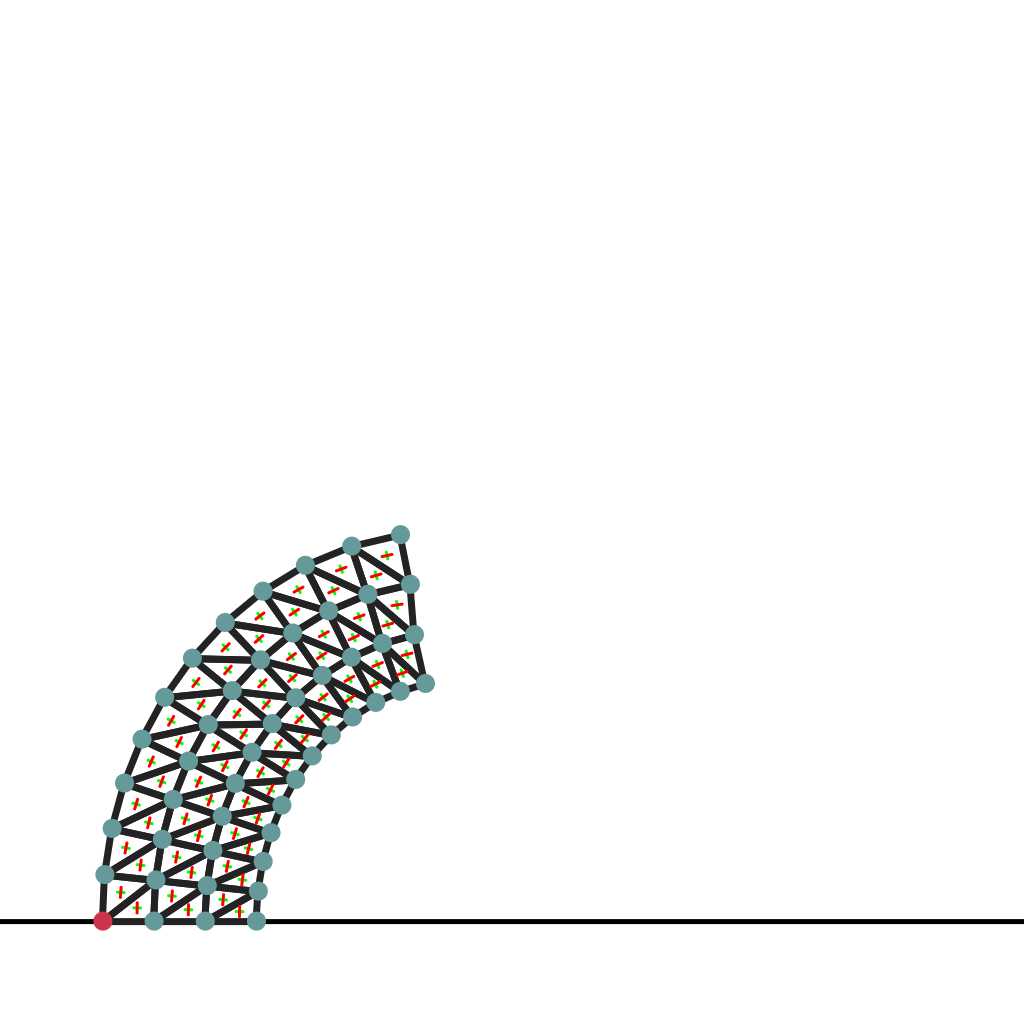}          \\
    \small{Template} & \small{Naive Warp} & \small{Proposed Warp}
    \end{tabular}
    \caption{Comparison of the warp methods. The first column is the template mesh with the actuation patterns (cross mark) in the canonical space. 
    The other columns are target meshes with the actuation patterns warped with different methods. Better to zoom in to see the orientation of the cross marks.}
    \label{fig:warp}
\end{figure}

\begin{figure}[!t] %
    \centering
    \begin{tabular}{
    P{0.3\linewidth}@{\hspace{3pt}}
    P{0.3\linewidth}@{\hspace{3pt}}
    P{0.3\linewidth}@{\hspace{3pt}}
    }
    \includegraphics[trim=0 0 100 0, clip, width=1\linewidth]{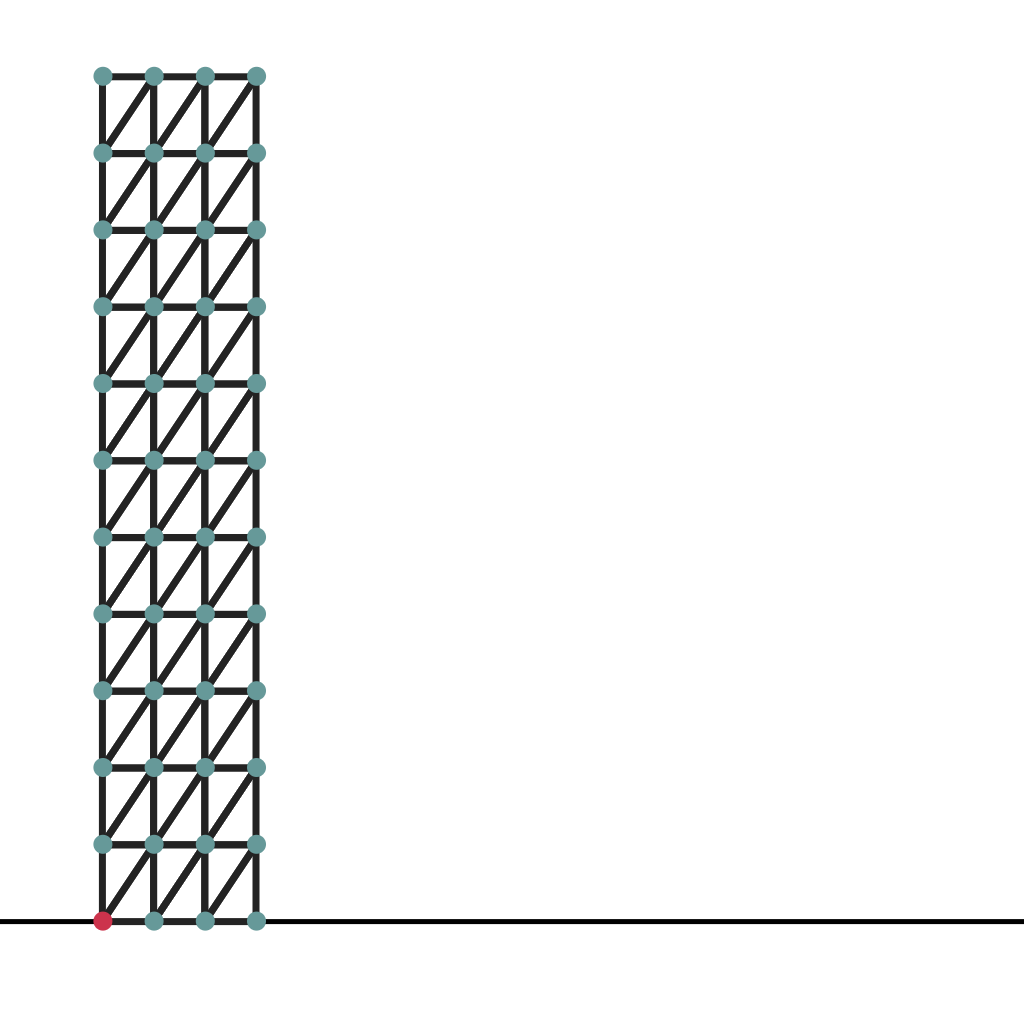}      &
    \includegraphics[trim=0 0 100 0, clip, width=1\linewidth]{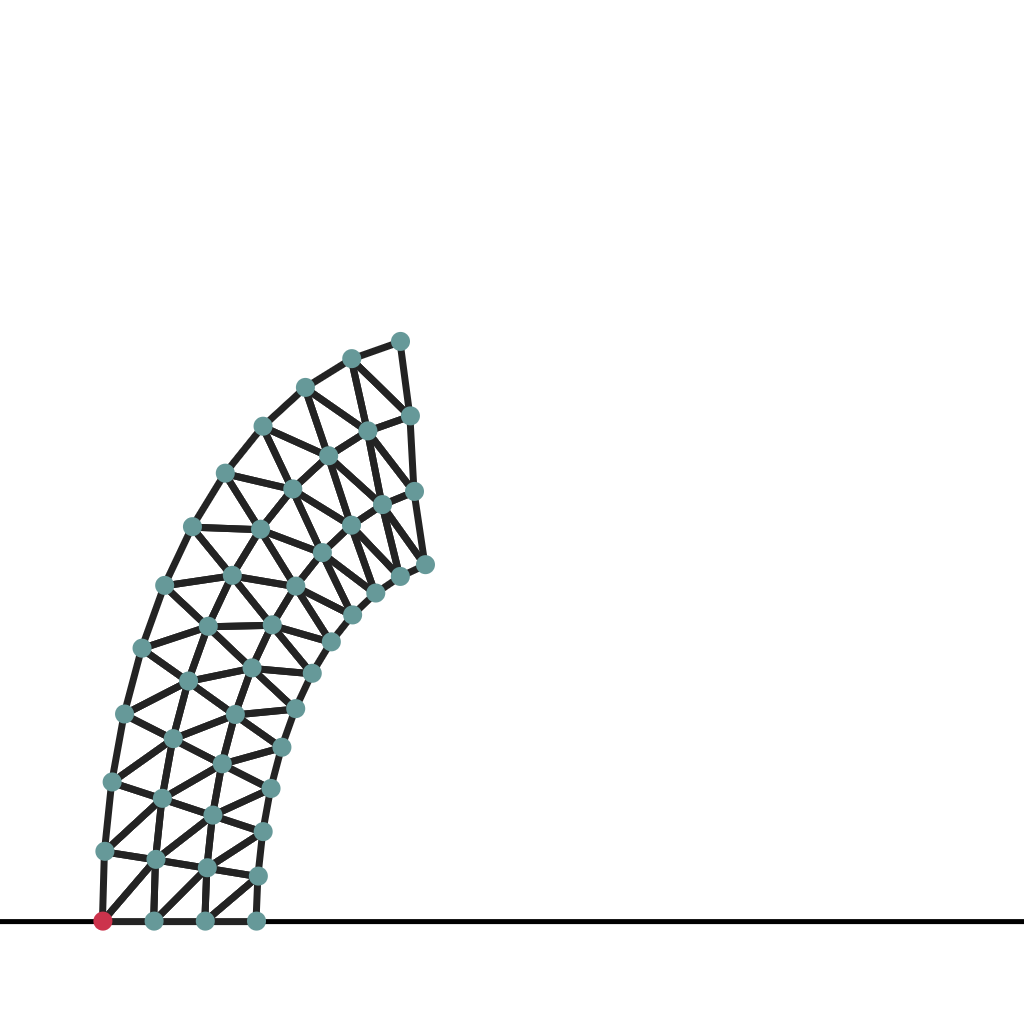}   &
    \includegraphics[trim=0 0 100 0, clip, width=1\linewidth]{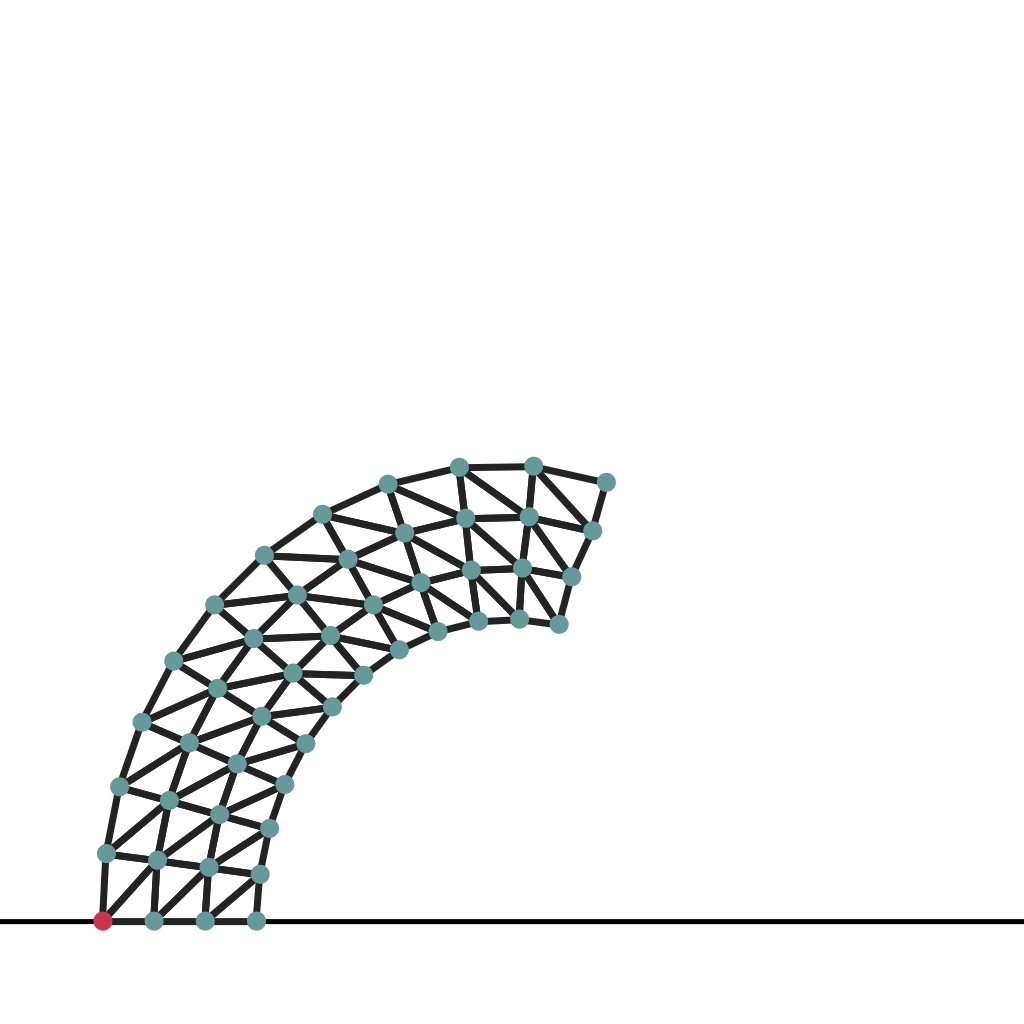}         \\
    \small{Reference} & \small{Naive Warp} & \small{Proposed Warp}
    \end{tabular}
    \caption{The first column is the reference mesh after deformation induced by the actuation.
    The other columns are the target meshes after deformation induced by the actuations warped with different methods.}
    \label{fig:warp_deformed}
\end{figure}

\begin{figure}[!t] %
    \centering
    \begin{tabular}{
    P{0.3\linewidth}@{\hspace{3pt}}
    P{0.3\linewidth}@{\hspace{3pt}}
    P{0.3\linewidth}@{\hspace{3pt}}
    }
    \includegraphics[trim=30 0 50 0, clip, width=1.2\linewidth]{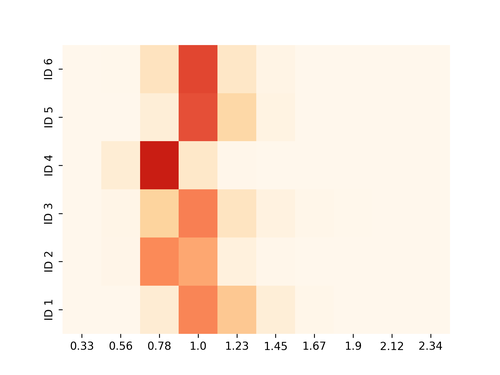}   &
    \includegraphics[trim=30 0 50 0, clip, width=1.2\linewidth]{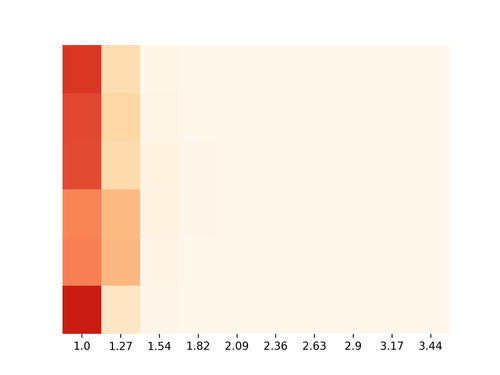}   &
    \includegraphics[trim=30 0 50 0, clip, width=1.2\linewidth]{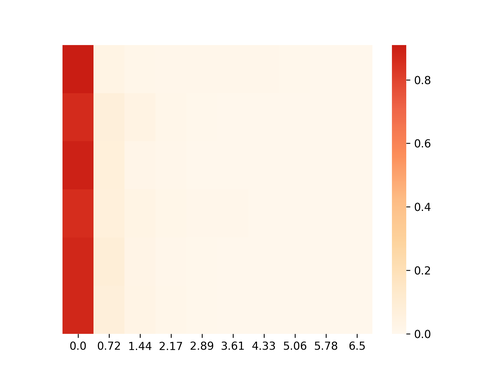}         \\
    \small{Determinant} & \small{Anisotropy} & \small{Error}   
    \end{tabular}
    \caption{The statistics of our mapping functions. The $y$-axis is identity number, and the $x$-axis is the value of the corresponding measurement.
    The color of the heatmap indicates the percentage of the sampled points that fall into the corresponding bin.
    The first column shows the heatmaps of determinant of the Jacobian matrix of the mapping functions,
    the second column shows the heatmaps of the ratio of the largest to the smallest singular values of the Jacobian matrix,
    and the third column shows the heatmaps of the vertex error.}
    \label{fig:mapping_fn}
\end{figure}

What should change is the actuation tensor, which could be decomposed into contractile directions and the magnitudes. 
Instead of directly warping the actuation tensor using $\mathcal{A}(\phi(\x))$, which is not semantically consistent as shown in Fig. \ref{fig:warp},
we couple it with the Jacobian of $\phi^{-1}$, \ie, $\frac{\partial{\phi^{-1}(\X)}}{\partial \X}$. 
Specifically, two things are taken into consideration.
First, the magnitudes should \emph{not} change along with $\frac{\partial{\phi^{-1}(\X)}}{\partial \X}$, 
otherwise there will be extra deformation induced for the rest shape of the target mesh.
Second, the contractile directions should correlate with $\frac{\partial{\phi^{-1}(\X)}}{\partial \X}$, e.g., to rotate consistently, 
thus preserving semantic meaning.
To achieve this goal, we could factorize out the rotational component $\mathbf{R}_{\phi^{-1}}$ of $\frac{\partial{\phi^{-1}(\X)}}{\partial \X}$,
then the warped actuation tensor is $\tilde{\mathcal{A}}(\x) = \mathbf{R}_{\phi^{-1}}\mathcal{A}(\phi(\x))\mathbf{R}_{\phi^{-1}}^{\top}$. 
The energy function for simulation is then 
\begin{equation}
    \label{eqn:warped}
    E = \int_{\domain{M}} \frac{1}{2}\left\|\mathbf{F(\x)}-\mathbf{R^{*}(\x)} \tilde{\mathcal{A}}(\x)\right\|_{F}^{2} dv
\end{equation}
Fig. \ref{fig:warp} and Fig. \ref{fig:warp_deformed} show the basic idea. 
In practice, we don't need to train two separate networks for $\phi$ and $\phi^{-1}$,
since we have the following implicit relation
\begin{equation}
    \label{eqn:implicit_relation}
    \frac{\partial{\phi^{-1}(\X)}}{\partial \X} = \frac{\partial{\phi(\x)}}{\partial \x}^{-1},
\end{equation}
therefore $\R_{\phi^{-1}}$ = $\R_{\phi}^{-1}$ = $\R_{\phi}^\top$. Putting this together,
we have $\tilde{\mathcal{A}}(\x) = \mathbf{R}_{\phi}^{\top}\mathcal{A}(\phi(\x))\mathbf{R}_{\phi}$. 
As shown in \figref{fig:networks}, 
the network architecture for $\phi$ is composed of 4 SIREN layers with the hyperparameter $\omega_0=5$, and one linear layer.
For simplicity, we train such a network for each identity. The training takes 100000 iterations, with a learning rate of $1e-4$ that starts to linearly decay to $0$ after 50000 iterations. 
The elastic regularization weight $\lambda_e$ in Eq. 2 in the main paper
is set to $10$. To evaluate this term, we sample $N_e$ vertices in total: the simulation vertices plus randomly sampled points (one point per element of the simulation mesh). 
Training takes about 1 hour.
After training, the necessary information for the warp operation is evaluated once and reused in the training of the multi-identity framework.
The statistics of our 6 mapping networks $\phi$ are shown in \figref{fig:mapping_fn}. Note that in addition to the vertex error of the explicit surface constraint, we also report the volumetric statistics of the Jacobian of $\phi$.
The volumetric statistics are generated by randomly sampled the same amount of points as in evaluating the elastic regularization term.

\subsection{Multi-identity Architecture and Training}
\begin{figure}[!b]
    \centering
    \includegraphics[trim=0 0 0 0, clip, width=0.9\linewidth]{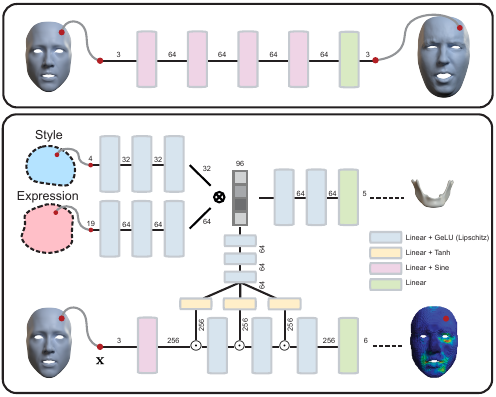}
    \caption{The architectures of our networks. The first row shows the architecture of our mapping network $\phi$. The bottom row shows the architecture of our multi-identity framework. The channel number for each layer is shown on the arrows.}
    \label{fig:networks}
\end{figure}

Our network architecture is shown in \figref{fig:networks}.
First, the input $19$-dimensional expression code and $4$-dimensional identity code are mapped into a higher dimensional space
via two tiny MLPs (akin to learned positional encoding), and subsequently get concatenated to result in the activation code $\z$.
$\z$ is the direct input to the generative transformation network $\network_{B}$, and also serves as the modulation input for the generative actuation network $\network_{A}$.
We design the activation functions shown in the figure for the following considerations.
\citeN{yang2022implicit} propose to use SIREN as the backbone of $\network_{A}$, 
which we find to be unstable, extremely sensitive to initialization, and prone to produce noisy results. Thus, we replace all the SIREN layers with GeLU layers except the first layer which serves as the learnable positional encoding.
Other positional encoding methods could also be applied here.
Since GeLU activation function is unbounded from above, we use the tanh activation function to bound the modulation input.
In order to add Lipschitz constraint, we augment each GeLU layer $i$ with a Lipschitz weight normalization layer \cite{liu2022learning} that has a trainable Lipschitz bound $c_{i}$.
For more details of the Lipschitz weight normalization layer, please refer to \citeN{liu2022learning}.

Following the two-stage training strategy inspired by \cite{yang2022implicit,srinivasan2021learning}, we commence with a plausible approximation of the actuation tensor field and jaw transformation for each target pose. This is based on passive muscle simulation \cite{srinivasan2021learning} and the tracking method delineated in \cite{Zoss2019}. This phase aids in warming up the training without the necessity for a differentiable simulator. In the second stage, we train the network with the integration of a collision-agnostic differentiable simulator.
For Eq. 7 in the main paper,
we assign values of $1e-3$ to $\lambda_{act}$ and $1e-6$ to $\lambda_{lip}$. We use the sampled actuation tensors used for simulation in $\mathcal{L}_{act}$. The initial stage runs for 400 epochs (roughly 16 hours), using a learning rate of $1e-4$ which linearly decays to zero after the 200th epoch. The second stage, lasting 20 epochs (approximately 30 hours), starts with a learning rate of $5e-5$ which begins a linear decay to zero right from the start.
The batch size for all stages is set to 6. Notably, in the second stage, the varying simulation meshes across identities render the computational graph identity-dependent, which precludes naive identity batching. To address this, we apply a distributed data parallel strategy and train our network on multiple GPUs. For the second stage, our framework takes around 6 seconds per iteration on average, covering both forward and backward passes.

\subsubsection{Simulation}
Recall that our simulation framework consists of three energy terms: shape targeting, bone attachment, and contact energies. 
The first two terms are based on Projective Dynamics \cite{Bouaziz2014ProjectiveSimulation}, where the local constraint can be generally represented as follows:
\begin{equation}
\label{eq:local}
E_i(\u) = \min_{\y_i} \frac{\omega_i}{2} ||\mathbf{G}_i \mathbf{S}_i \u - \mathbf{B}_i \y_i||_F^2 \;\; \text{s.t.} \;\; C_i(\y_i) = 0,
\end{equation}
where $\u$ denotes the simulation vertices, $\omega_i$ is a weight coefficient, and $\y_i$ an auxiliary variable, embodying the target position.
$\mathbf{S}_i$ is a selection matrix choosing DOFs involved in $E_i$.
$\mathbf{G}_i$ and $\mathbf{B}_i$ are designed to facilitate the distance measure.
For the shape targeting energy, 
$\mathbf{G}_i$ maps $\u$ to the deformation gradient $\F_i$. 
$\mathbf{B}_i$ comes from the input actuation tensor $\A_i$ and $\y_i$ denotes the rotation matrix, 
projected from $\F_i \A_i$.
For the bone attachment energy, $\mathbf{G}_i$ extracts the embedded bone vertex from $\u$, 
while $\mathbf{B}_i$ becomes an identity matrix and $\y_i$ is directly the given target position.
The total energy $E(\u)$ is the sum of all these local constraints.
After converging to a local minimum, we can calculate the sensitivity matrices for the input variables of interest with implicit differentiation.
For example, the sensitivity matrix of $\u$ with respect to $\A_i$ is given by:
\begin{equation}
\label{eq:sensitivity}
\frac{\partial \u}{\partial \A_i} = - \left(\nabla^2 E\right)^{-1} \frac{\partial \nabla E}{\partial \A_i}.
\end{equation}

For collision modeling, we employ the IPC model \cite{li2020incremental}, which utilizes the incremental barrier energy $B(\u)$. We set the distance threshold $\hat{d}$ to $0.001l$, where $l$ denotes the diameter of the chosen identity. With respect to differentiable simulation, the calculation of the sensitivity matrices needs adjustment. For instance, the sensitivity matrix of $\u$ in relation to $\A_i$ is given by:
\begin{equation}
\label{eq:sensitivity2}
\frac{\partial \u}{\partial \A_i} = - \left(\nabla^2 E + \nabla^2 B\right)^{-1} \frac{\partial \nabla_\u E}{\partial \A_i}.
\end{equation}
Note that incorporating the collision model into the simulation will increase the computational cost. 
For each frame, simulating from the rest shape takes around 30 seconds on average.
However, we find that during animation, using the previous frame as the initial state can significantly reduce the simulation time to around 6-10 seconds per frame on average.

\subsection{Other Experimental Results}

\paragraph{Comparison with the displacement network.}
We demonstrate the superiority of our physics-based model by contrasting it with a displacement variant that disregards physics and directly regresses the displacement field. 
In order to maintain parity, the displacement field is also learned in the canonical space, employing the same Lipschitz regularization and geometric loss function as our actuation network. We utilize a similar network architecture, simply adjusting the dimension of the final layer to three.

\begin{figure}[!htb]
    \centering
    \includegraphics[width=\linewidth]{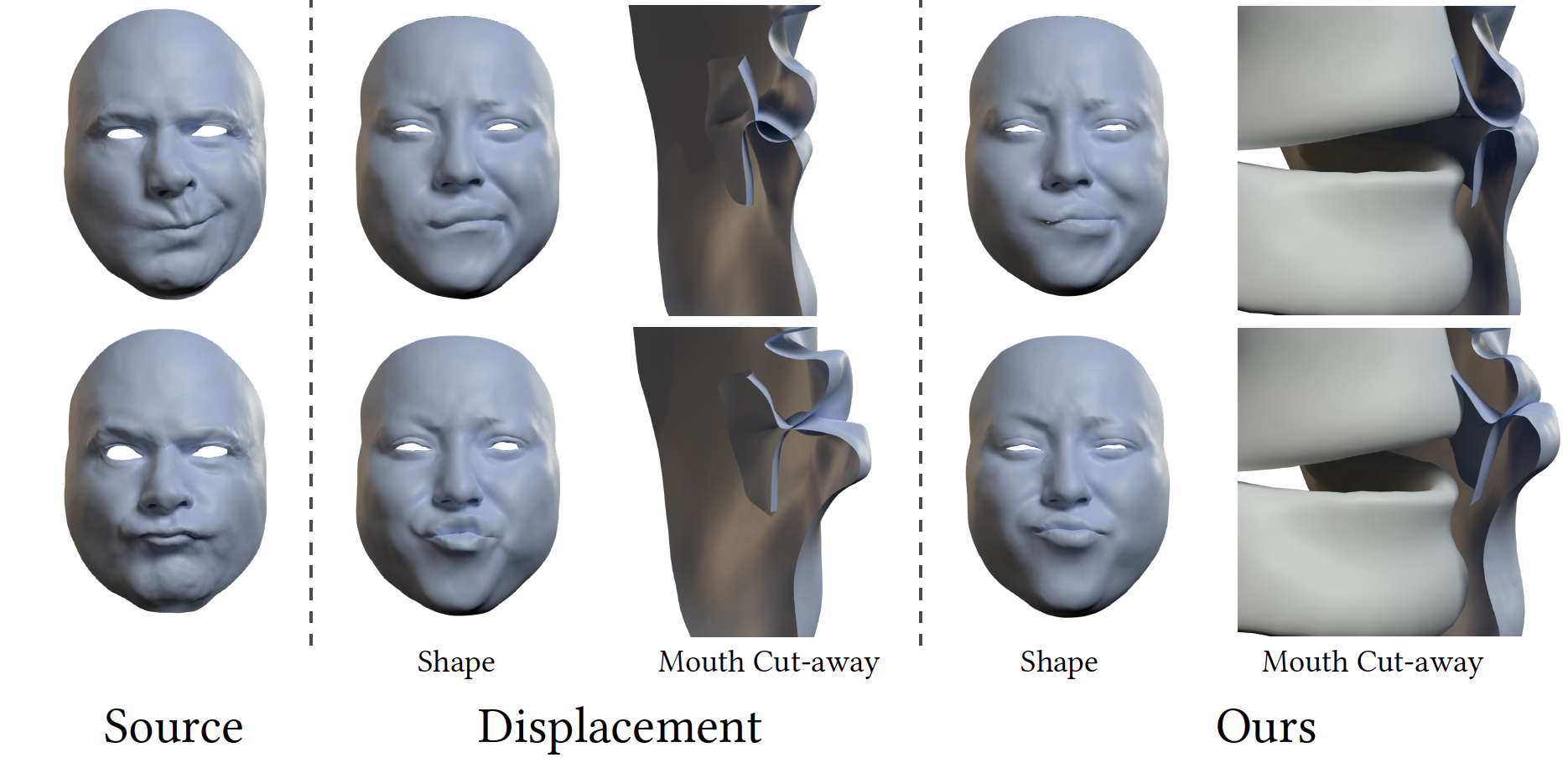}
    \caption{Comparison between our displacement network and our actuation network in terms of retargeting.}
    \label{fig:comparedisplacement}
\end{figure}

As shown in \figref{fig:comparedisplacement}, despite the displacement network's inability to manage collision, it also suffers from problems like sudden shape distortion in the lip region, further emphasizing the benefits of our physics-based approach.

\begin{figure}[!htb]
    \centering
    \includegraphics[width=1.0\linewidth]{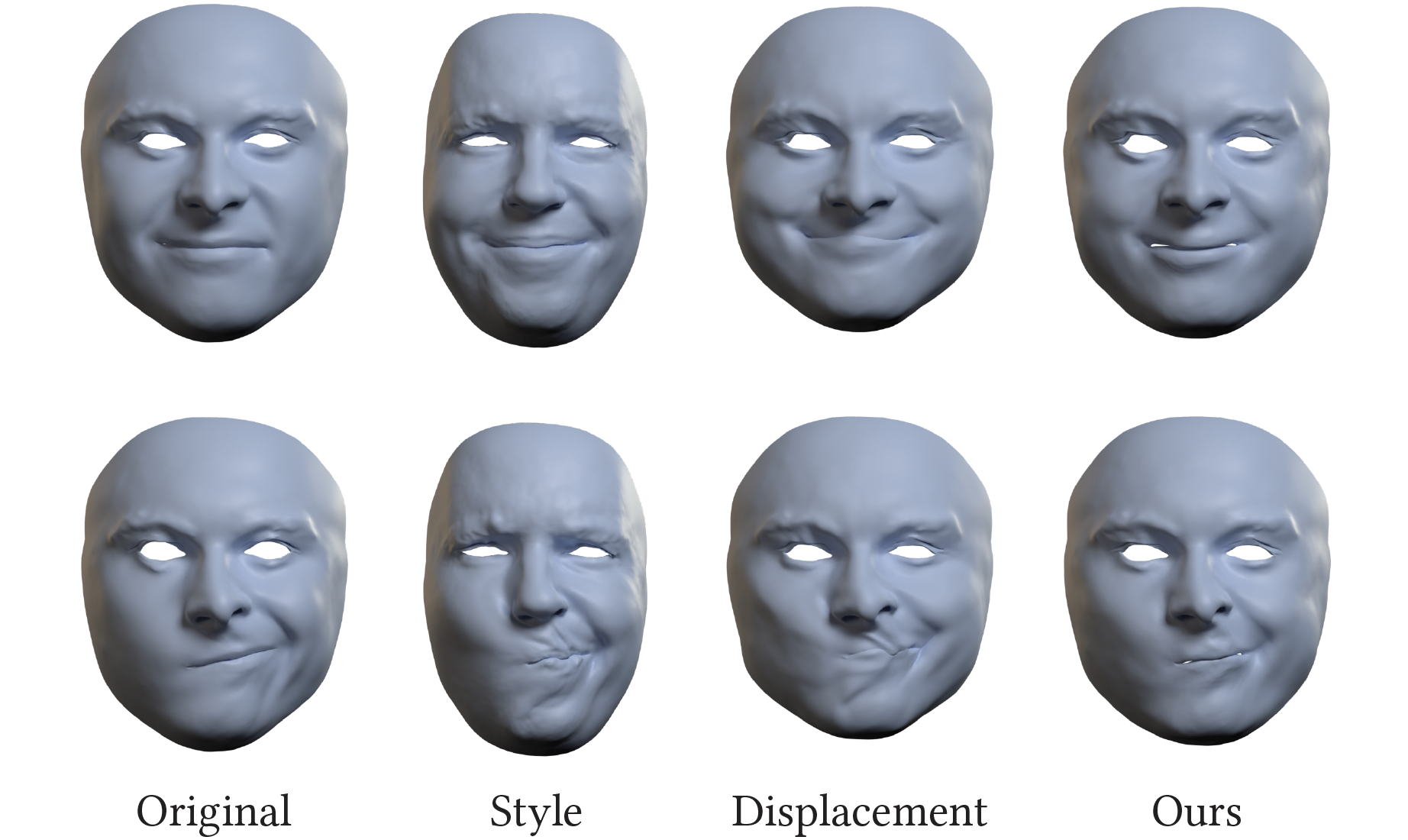}
    \caption{Style transfer comparison between our model and the model without canonical space (Model-N). Note how the actuation field is inconsistent across the identities in Model-N for the same style.}
    \label{fig:styletransfercomparisondisplacement}
    \vspace{-0.5cm}
\end{figure}

We also compare the displacement network with our physics-based model in terms of style transfer task.
As shown in \figref{fig:styletransfercomparisondisplacement}, our physics-based model can better preserve the identity and expression of the source face, while the displacement network suffers from severe volume change and lip penetration in the lip region.


\end{document}